\documentclass[10pt,journal,compsoc]{IEEEtran}
\ifCLASSOPTIONcompsoc
  \usepackage[nocompress]{cite}
\else
  \usepackage{cite}
\fi
\ifCLASSINFOpdf
\else
\fi

\usepackage{epsfig}
\usepackage{graphicx}
\usepackage{amsmath}
\usepackage{amssymb}
\usepackage{enumitem}
\usepackage{ragged2e}

\usepackage[pagebackref=true,breaklinks=true,letterpaper=true,colorlinks,bookmarks=false,urlcolor=red,citecolor=cyan]{hyperref}
\usepackage{gensymb,graphics,subfigure,inputenc}
\usepackage[linesnumbered,algo2e,boxed]{algorithm2e}
\graphicspath{{Figure/}}
\usepackage[figuresright]{rotating}
\usepackage{amsmath,amssymb,textcomp,subfigure,multirow,upgreek}
\usepackage{booktabs}
\usepackage{color,mdwlist}

\newcommand{\hao}[1]{\textcolor{black}{#1}}

\usepackage{graphicx,fontawesome}
\graphicspath{{Figures/}}

\usepackage{arydshln}

\begin{document}
%
\title{Spatial-Temporal Graph Mamba for Music-Guided Dance Video Synthesis}
\author{\IEEEauthorblockN{Hao Tang \quad
        Ling Shao \quad
        Zhenyu Zhang \quad
        Luc Van Gool	\quad
	Nicu Sebe 
	}
	\thanks{\par Hao Tang is with the State Key Laboratory of Multimedia Information Processing,
School of Computer Science, Peking University, China. Email: haotang@pku.edu.cn
		\par 
        Ling Shao is with the UCAS-Terminus AI Lab, University of Chinese Academy of Sciences, Beijing 100049, China \protect
        \par Zhenyu Zhang is with the School of Intelligent Science and Technology, Nanjing University, China. \protect
    \par Luc Van Gool is with the Department of Information Technology and
    Electrical Engineering, ETH Zurich, Switzerland, with the Department of Electrical Engineering, KU Leuven, Belgium, and with INSAIT, Sofia Un., Bulgaria. \protect
    \par Nicu Sebe is with Department of Information Engineering and Computer Science (DISI), University of Trento, Italy.  \protect
		\par Corresponding authors: Hao Tang and Ling Shao}
}

%
%

\markboth{IEEE Transactions on Pattern Analysis and Machine Intelligence}%
{Shell \MakeLowercase{\textit{et al.}}: Bare Demo of IEEEtran.cls for Computer Society Journals}
%



\IEEEtitleabstractindextext{%
\justify

\begin{abstract}
We propose a novel spatial-temporal graph Mamba (STG-Mamba) for the music-guided dance video synthesis task, i.e., to translate the input music to a dance video. STG-Mamba consists of two translation mappings: music-to-skeleton translation and \hao{skeleton-to-video} translation. In the music-to-skeleton translation, we introduce a novel spatial-temporal graph Mamba (STGM) block to effectively construct skeleton sequences from the input music, capturing dependencies between joints in both the spatial and temporal dimensions. For the \hao{skeleton-to-video} translation, we propose a novel self-supervised regularization network to translate the generated skeletons, along with a conditional image, into a dance video. Lastly, we collect a new \hao{skeleton-to-video} translation dataset from the Internet, containing 54,944 video clips. Extensive experiments demonstrate that STG-Mamba achieves significantly better results than existing methods.
\end{abstract}

\begin{IEEEkeywords}
Spatial-Temporal Graph, Mamba, Music-Guided Dance Video Synthesis
\end{IEEEkeywords}}

\maketitle

\IEEEdisplaynontitleabstractindextext

%
\IEEEpeerreviewmaketitle


%
%
%
%

\section{Introduction}
\begin{figure}[tbp] \small
	\centering
	\includegraphics[width=1\linewidth, height=0.04\linewidth]{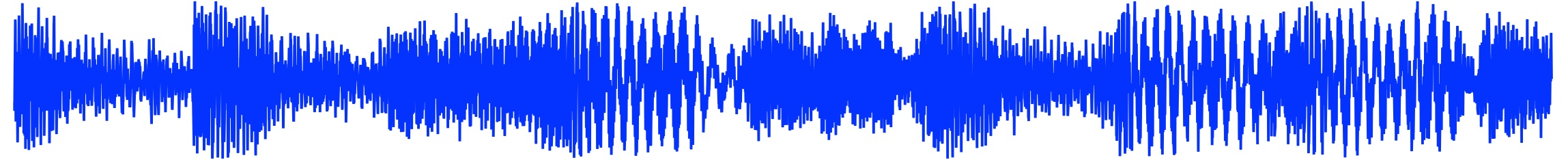}
	\includegraphics[width=1\linewidth]{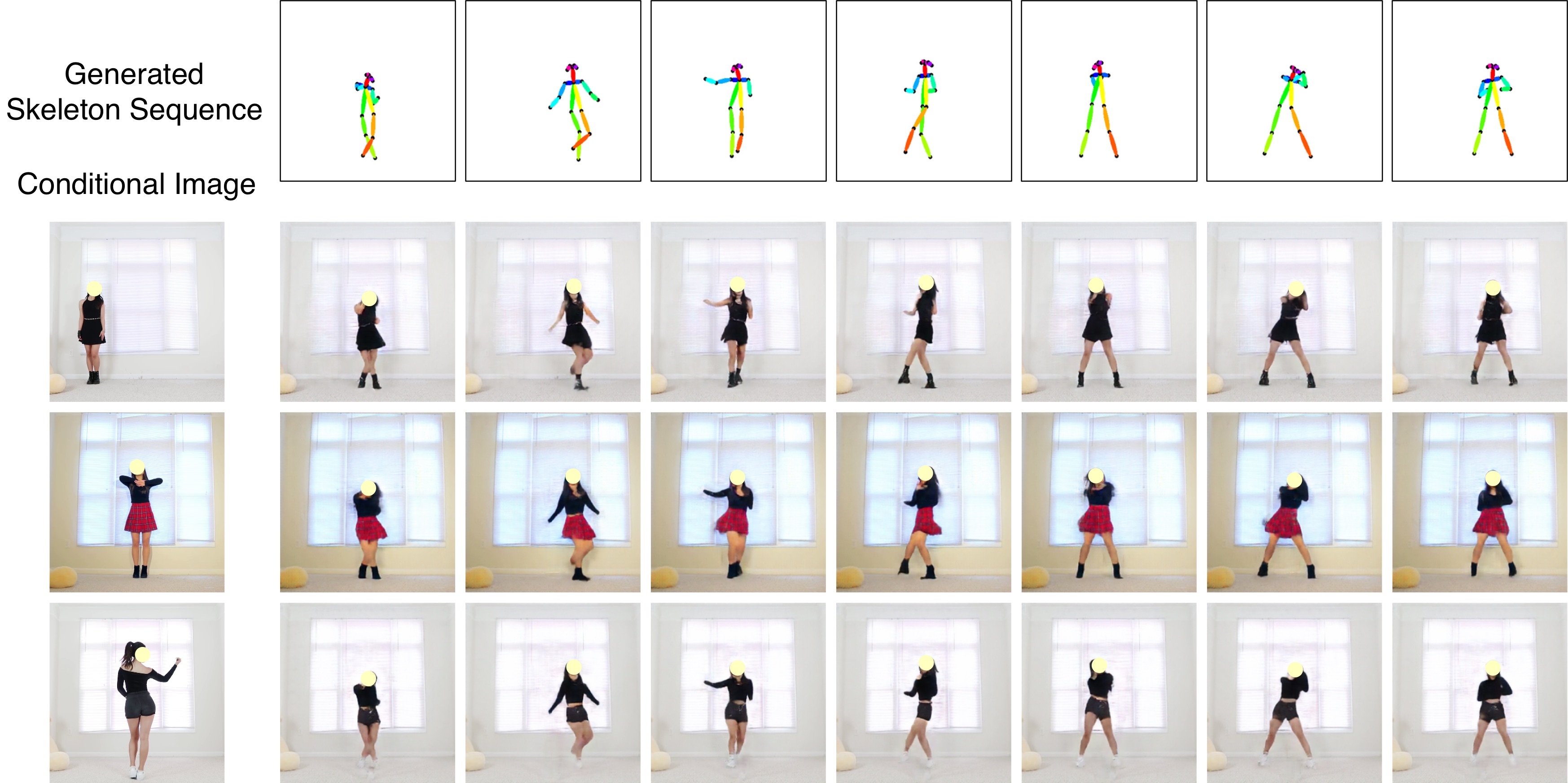}
	\caption{Given an input music, the proposed STG-Mamba first generates a skeleton sequence and then generates realistic dance videos using different conditional images.}
	\label{fig:sota2_10}
\end{figure}

\begin{figure*}[!t] \small
	\centering
	\includegraphics[width=0.9\linewidth]{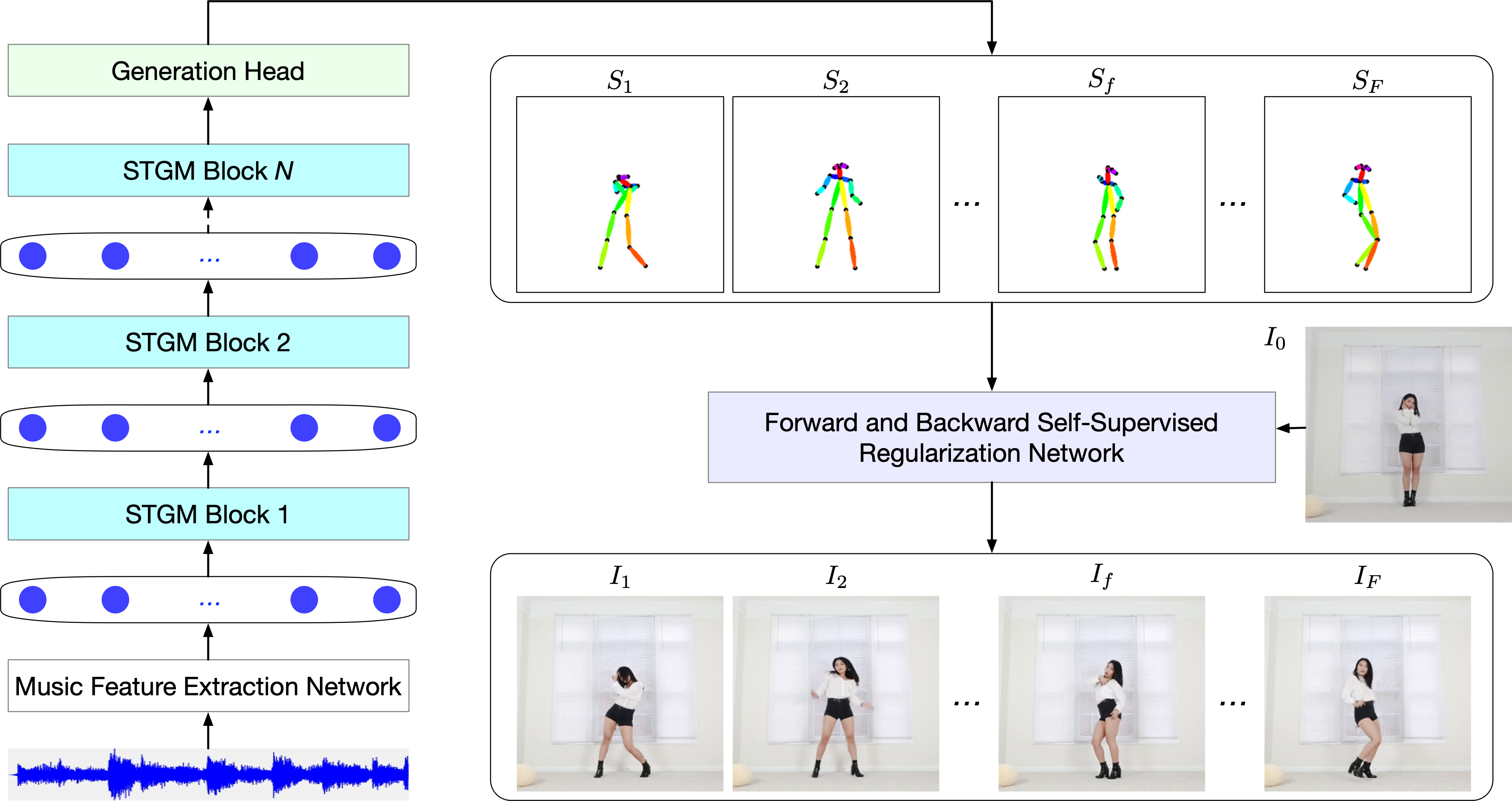}
	\caption{Overview of the proposed STG-Mamba, which consists of two translation mappings. The music-to-skeleton translation consists of a music feature extraction network, a series of spatial-temporal graph Mamba (STGM) blocks, and a generation head. The proposed STGM block aims to capture the dependencies between human joints in spatial and temporal dimensions. 
	The \hao{skeleton-to-video} translation contains a novel forward and backward self-supervised regularization network to translate the generated skeleton sequence and a conditional image $I_0$ to a photorealistic person dancing video.}
	\label{fig:method}
\end{figure*}

In this paper, we focus on the music-guided dance video synthesis task, i.e., translating input music to a photorealistic dance video, as depicted in Figure~\ref{fig:sota2_10}.
However, music/audio data tend to be less structured, making it highly challenging to model its relationships with visual/video data. This in turn makes generating realistic dance videos from input music an unsolved problem. 
We decompose the synthesis task into two translation mappings, as shown in Figure~\ref{fig:method}: 
1) music-to-skeleton translation, i.e., we generate skeleton sequences from the input music to predict the dance moves, and 2) \hao{skeleton-to-video} translation, i.e., we generate a dance video conditioned on the generated skeleton sequences, as well as a conditional image.
In this way, we simultaneously model the motion information from the generated skeleton sequences and the appearance content (e.g., person identity) from the conditional image, leading to a final realistic dance video.

Existing music-to-skeleton translation methods are mainly based on classical convolutional and recurrent neural networks \cite{lee2019dancing,kim2006making,guo2020action2motion,kao2020temporally,shlizerman2018audio,ginosar2019learning,ye2020choreonet,alemi2017groovenet,shiratori2006dancing,ahn2020generative,tang2018dance}.
For instance, Tang et al. \cite{tang2018dance} use an LSTM-autoencoder model to generate dance pose skeletons, while Lee et al. \cite{lee2019dancing} introduced a synthesis-by-analysis learning framework to generate dance from music. 
However, these networks suffer from training and variability issues due to the non-Euclidean geometry of the motion manifold structure.
To address these issues, some methods, such as \cite{ren2020self,chen2021choreomaster} 
employ graph convolutional networks (GCNs) \cite{li2019actional,si2019attention,yan2018spatial} for skeleton generation from input music. For instance, Ren et al.~\cite{ren2020self} presented a pose perceptual loss that relies on a pre-trained GCN network to match intermediate features, assuming that the GCN contains high-level spatial and structural information of the human skeleton.
However, modeling the skeleton as a graph only during loss calculation is insufficient for capturing the dependencies and correlations between human joints and thus potentially results in distorted actions.
In biomechanics, the movement and positioning of one joint is often dependent on and correlated with other joints. For instance, bending the elbow typically involves not only the elbow joint but also muscles and joints in the shoulder, wrist, and even fingers.
A graph-based approach can model these dependencies by representing each joint as a node and the connections between joints as edges. However, if this is only done during the loss calculation, the dependencies and correlations between joints might not be fully incorporated into the learned model.
As a result, the actions or movements generated by the model might appear unnatural or distorted because the underlying joint interdependencies aren't properly modeled throughout the learning process. Thus, it is more effective to model the skeleton as a graph not just during loss calculation but consistently throughout the entire process of learning and generating movements.

In addition, there are some methods \cite{siyao2022bailando,li2022danceformer} to solve this problem using transformers \cite{vaswani2017attention}.
For example, Li et al. \cite{li2022danceformer} proposed DanceFormer to learn music-conditioned dance generation by modeling it as key pose generation and parametric motion curve regression. 
Despite their remarkable success, transformer models come with significant computational inefficiencies, particularly evident when handling long sequences. Unlike RNNs or CNNs, transformers exhibit quadratic time complexity relative to sequence length, rendering them less scalable for processing extensive input. This inefficiency poses challenges in real-world applications where processing large-scale data is essential. Moreover, the sheer size and complexity of transformer architectures require substantial computational resources, hindering their deployment in resource-constrained environments. Additionally, transformers' self-attention mechanism, while powerful for capturing global dependencies, contributes to this computational burden, as it requires pairwise comparisons across all tokens in the sequence.

To address these limitations, in this paper, 
we explicitly treat each human joint as a single node in a graph during the early generation stage instead of only during loss calculation.
Moreover, to better capture dependencies between joints/nodes both spatially and temporally, we propose a novel spatial-temporal graph Mamba (STGM) block, as shown in Figure~\ref{fig:block}.
The core components of the STGM block are three state space models (SSMs), i.e., \hao{spatial graph SSM (SG-SSM)}, \hao{temporal graph backward SSM (TGB-SSM)}, and \hao{temporal graph forward SSM (TGF-SSM)}, since SSMs are more computationally efficient for modeling sequential data, and require fewer computational resources, especially when dealing with long sequences \cite{gu2023mamba}.
Specifically, SG-SSM aims to understand self-frame spatial interactions between different body joints (Figure~\ref{fig:block}, top right). Meanwhile, TGB-SSM and TGF-SSM aim to model cross-frame temporal correlations from both directions (Figure~\ref{fig:block}, right middle and bottom).
Through a series of STGM blocks (see Figure \ref{fig:method}), we can impose tighter constraints to obtain more accurate joint predictions.

To translate the generated skeleton sequence into a photorealistic dance video, existing methods always directly adopt a state-of-the-art motion transfer model \cite{chan2019everybody,liu2019liquid,wang2018video,zhou2019dance,wang2019few}.
For example, Ren et al.~\cite{ren2020self} directly used Pix2pixHD \cite{wang2018high} to transfer the generated skeletons to the target dance video.
Lee et al. \cite{lee2019dancing} adopted Vid2Vid \cite{wang2018video} to convert skeleton sequences to videos.
Despite their success, we still observe that existing methods (e.g., \cite{wang2018high,wang2018video}) produce unsatisfactory moves and visual artifacts in the generated videos, which we believe are mainly caused by the inconsistent generation order.

To address this generation disorder, we propose a novel forward and backward self-supervised regularization network to generate realistic dance videos conditioned on the generated skeleton sequence and a conditional image.
This regularization network generates videos using three strategies, i.e., baseline generation, forward generation, and backward generation (see Figure~\ref{fig:stage2}).
Additionally, we introduce self-supervised regularizations for the forward and backward passes to facilitate video generation and enhance training stability.

\hao{As an additional contribution, we collected a new skeleton-to-video translation dataset from the Internet, consisting of 54,944 video clips. 
To the best of our knowledge, this dataset is the largest skeleton-to-dance-video dataset to date, containing 54,944 video sequences with diverse dance styles (ballet, popping, and K-pop). Compared to existing datasets, ours provides more diversity in terms of dancer gender distribution (62\% female, 38\% male) and environmental settings (76\% indoor, 24\% outdoor). This diversity enhances the model's generalization ability to different dance forms and real-world scenarios.}

Overall, our contributions are as follows:
\begin{itemize}[leftmargin=*]
	\item We propose a novel STG-Mamba for the challenging music-guided dance video synthesis task, which consists of two generation mappings, i.e., music-to-skeleton translation and skeleton-to-video translation.
	\item We propose a novel STGM block to explicitly model dependencies across joints both spatially and temporally at the same time. 
	\item We design a new self-supervised regularization network to enhance video generation in both forward and backward directions.
    \item \hao{We collected a new skeleton-to-dance-video translation dataset from the Internet, consisting of 54,944 video clips.}
    \item  Extensive experiments verify that the proposed method produces results that are considerably better than the existing methods. 
\end{itemize}

\section{Related Work}

\noindent \textbf{Music-Guided Dance Video Synthesis} aims to generate dance videos from input music.
Traditional methods \cite{shiratori2006dancing,ofli2008audio,lee2013music,fan2011example} usually use statistical models to perform this task.
Recently, several works have used convolutional \cite{alemi2017groovenet,lee2019dancing,ferreira2021learning,zhuang2022music2dance}, recurrent neural networks \cite{tang2018dance}, and transformers \cite{li2021ai,li2022danceformer,li2020learning} to learn the mapping from the input music to the output dance video, harnessing the power of deep learning.
For instance, Tang et al. \cite{tang2018dance} introduced a music-guided dance choreography synthesis method using an LSTM-autoencoder model to learn a mapping between acoustic and motion features.

\noindent \textbf{Graph-Based Models} have been shown to be efficient for many tasks, including skeleton-based action recognition \cite{li2019actional,yan2018spatial,zhao2022pb}, semi-supervised classification \cite{kipf2016semi}, crowd counting \cite{chen2020relevant}, node classification \cite{velivckovic2017graph}, text classification \cite{yao2019graph}, anomaly detection \cite{zhong2019graph}, face clustering \cite{yang2019learning}, relation extraction \cite{zhang2019attention}, facial image synthesis \cite{tang2023bipartite}, image segmentation \cite{ling2019fast}, person image generation \cite{tang2020bipartite}, traffic forecasting \cite{guo2019attention}, human pose estimation \cite{li2025graphmlp},  architectural layout generation \cite{tang2024graph,tang2023graph}, interaction generation \cite{chopin2024bipartite}, and scene graph generation \cite{chen2019knowledge}.
Unlike previous methods, we employ graph-based models to solve a new task, i.e., music-to-skeleton translation.
Ren et al.~\cite{ren2020self} presented a pose perceptual loss based on a pre-trained GCN network to match the intermediate features and generate a realistic skeleton sequence from input music.
However, during our experiments, we discover that only modeling the skeleton as a graph during the loss calculation is not sufficient to capture long-range dependencies and correlations between human joints, which potentially leads to distorted actions.
This paper considers each human joint as a node in a graph during generation and further designs a novel graph-based Mamba block to capture long-range dependencies between joints in both the spatial and temporal dimensions.

\noindent \textbf{State Space Models} (SSMs) have seen widespread adoption across various domains for modeling sequential data with latent states and observed measurements. Notable advances in this area include the introduction of innovative models such as \cite{gu2021efficiently,smith2022simplified,fu2022hungry,mehta2022long,gu2023mamba}.
For example, 
Gu et al. \cite{gu2021efficiently} proposed the structured state-space sequence (S4), which offers linear scalability with sequence length, providing a compelling alternative to CNNs or transformers for capturing long-range dependencies. 
More recently, Gu and Dao \cite{gu2023mamba} introduced the Mamba model, featuring a data-dependent SSM layer, which surpassed transformers in performance across various sizes in extensive real-world datasets while maintaining linear scalability with sequence length. These advances collectively highlight the versatility and effectiveness of SSMs in modeling sequential data.

SSMs have also found extensive application in visual domains, including video classification, understanding, and segmentation \cite{islam2022long, nguyen2022s4nd,wang2023selective,yang2024vivim}, as well as movie scene detection \cite{islam2023efficient}, text-to-motion generation \cite{zhang2025motion}, image compression \cite{zeng2025mambaic}, and image segmentation \cite{ma2024u,ruan2024vm,liu2024swin,ma2024u,xing2024segmamba}. Although the existing literature has explored the use of SSMs in various visual tasks, we propose a novel graph-based SSM model tailored specifically for the challenging task of music-guided dance video synthesis. Unlike previous approaches that focus primarily on visual applications, our model leverages graph-based representations to effectively capture the complex relationships between music and dance movements, thereby enabling more accurate and expressive video synthesis.

\noindent \textbf{Self-Supervised Learning} has shown effectiveness in many tasks, including semantic segmentation \cite{wang2020self,pan2020unsupervised}, depth estimation \cite{tan2020self}, object recognition \cite{zhou2020look}, representation learning \cite{goyal2019scaling,jenni2020steering}, audio-visual speaker tracking \cite{li2022multi}, image generation \cite{tewari2020stylerig,choi2020inference}, action recognition \cite{jenni2020video}, optical flow estimation\cite{liu2020learning}, and common sense reasoning \cite{klein2020contrastive}.
Unlike these methods, in this paper, we adopt self-supervised learning to tackle the \hao{skeleton-to-video} translation task.
Specifically, we propose a novel self-supervised regularization network to generate realistic dance videos in both the forward and backward passes.
Moreover, the self-supervised regularizations for these two passes are proposed to facilitate the video generation process, and thus enhance the training stability.
\section{The Proposed STG-Mamba}

\begin{figure*}[!t]
	\centering
	\includegraphics[width=0.9\linewidth]{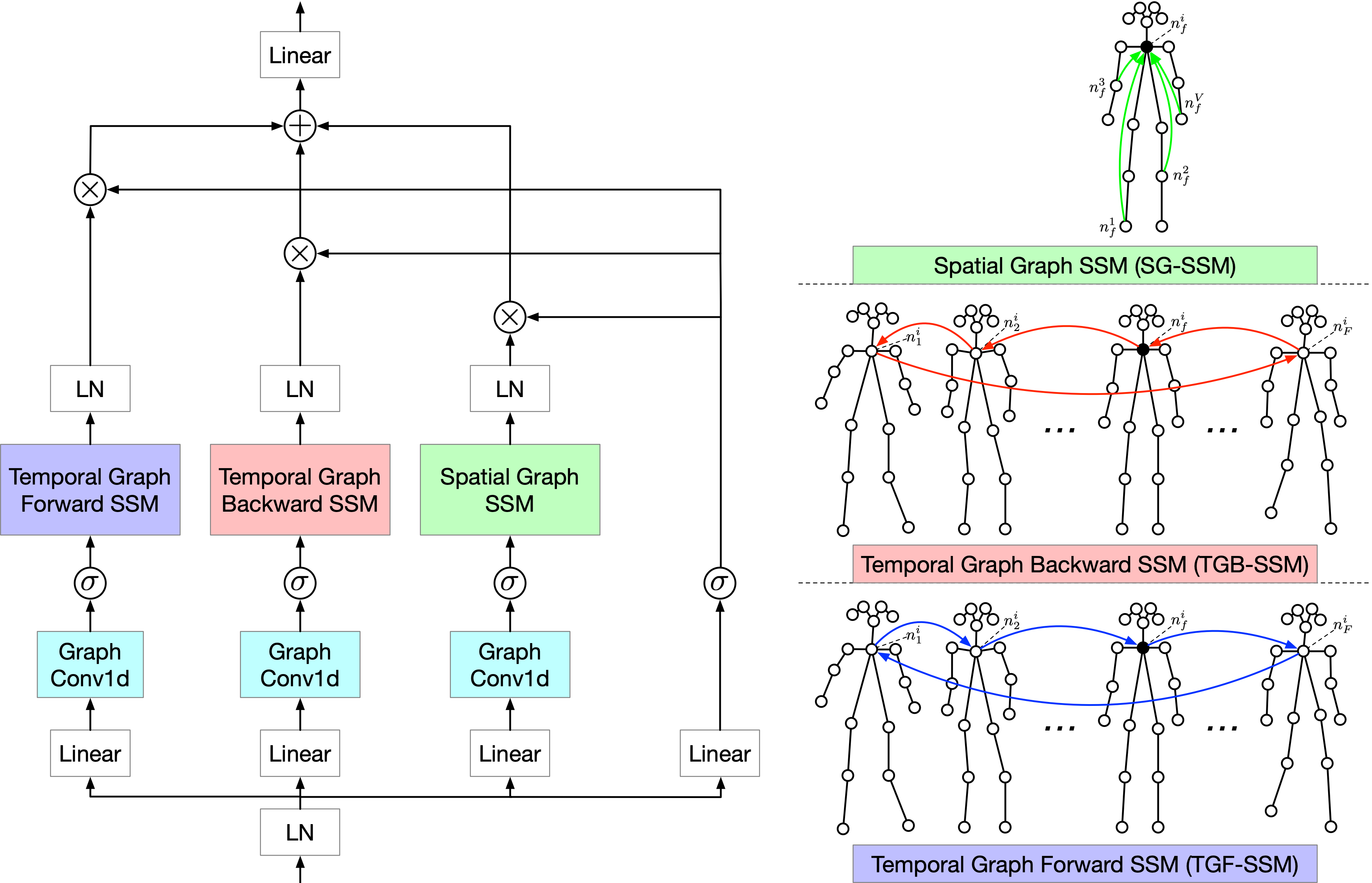}
	\caption{\textbf{Left:} Overview of the proposed spatial-temporal graph Mamba (STGM) block, which contains a temporal graph forward SSM (TGF-SSM), a temporal graph backward SSM (TGB-SSM), and a spatial graph SSM (SG-SSM). \textbf{Right:} Illustration of how these three SSMs capture spatio-temporal dependences.}
	\label{fig:block}
\end{figure*}

We start by introducing the details of the proposed STG-Mamba for music-guided dance video synthesis.
\hao{While Mamba was originally designed for very long sequences (over 2,048 elements), the core advantage of Mamba lies in its ability to capture both local and long-range dependencies efficiently. Although the skeleton sequences in dance generation are shorter, modeling the complex spatial–temporal relationships among joints remains challenging. The Mamba structure allows us to robustly capture these dependencies without being limited by the sequence length.}

The proposed STG-Mamba consists of two translation mappings (i.e., music-to-skeleton translation and \hao{skeleton-to-video} translation). 
An illustration of STG-Mamba is shown in Figure~\ref{fig:method}.
The first mapping contains three main parts, i.e., a music feature extraction network extracting the style and beat features from the input music, a series of STGM blocks modeling the long-range correlations between human joints spatially and temporally, and a generation head to output the skeleton sequence.
In the second translation, we take the generated skeleton sequence and a conditional image $I_0$ as input, aiming to generate a photorealistic dance video.
To this end, we introduce a new self-supervised regularization network to enhance the video generation process in both forward and backward generation directions.

\subsection{Music-to-Skeleton Translation}
\noindent \textbf{Overview.}
In this translation, the model aims to learn a mapping from the input music $M$, with a \hao{sampling} rate $r$, to a joint location vector sequence $S$, i.e., $[M {\in} \mathbb{R}^{Fr}, z] {\rightarrow} S {\in} \mathbb{R}^{F{\times}2V}$, where $F$ is the total number of video frames, and $V$ denotes the number of human joints in each frame, which are represented by a set of 2D coordinates, $z$ is a noise vector.

\begin{figure*}[t]
	\centering
	\includegraphics[width=0.32\textwidth]{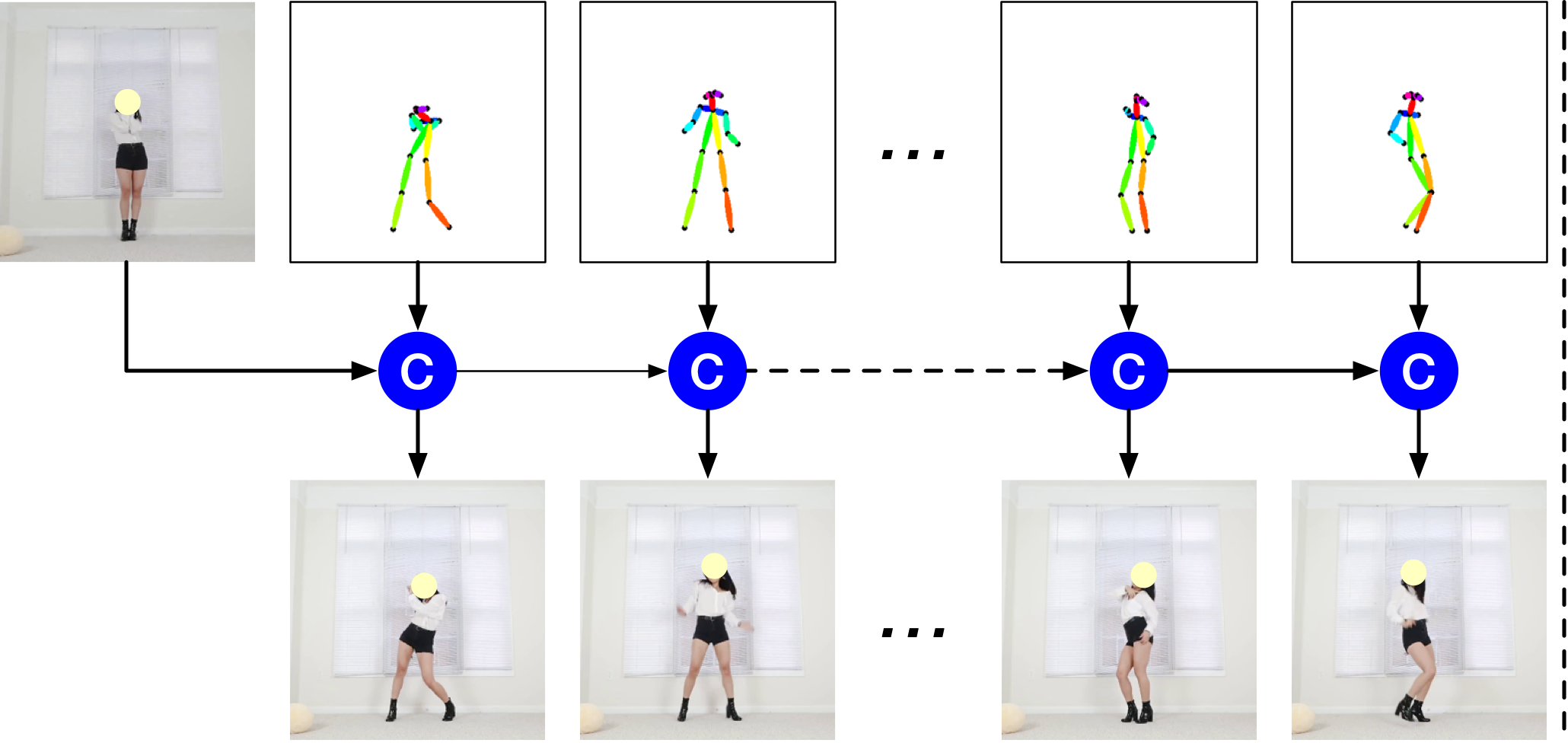}
	\includegraphics[width=0.32\textwidth]{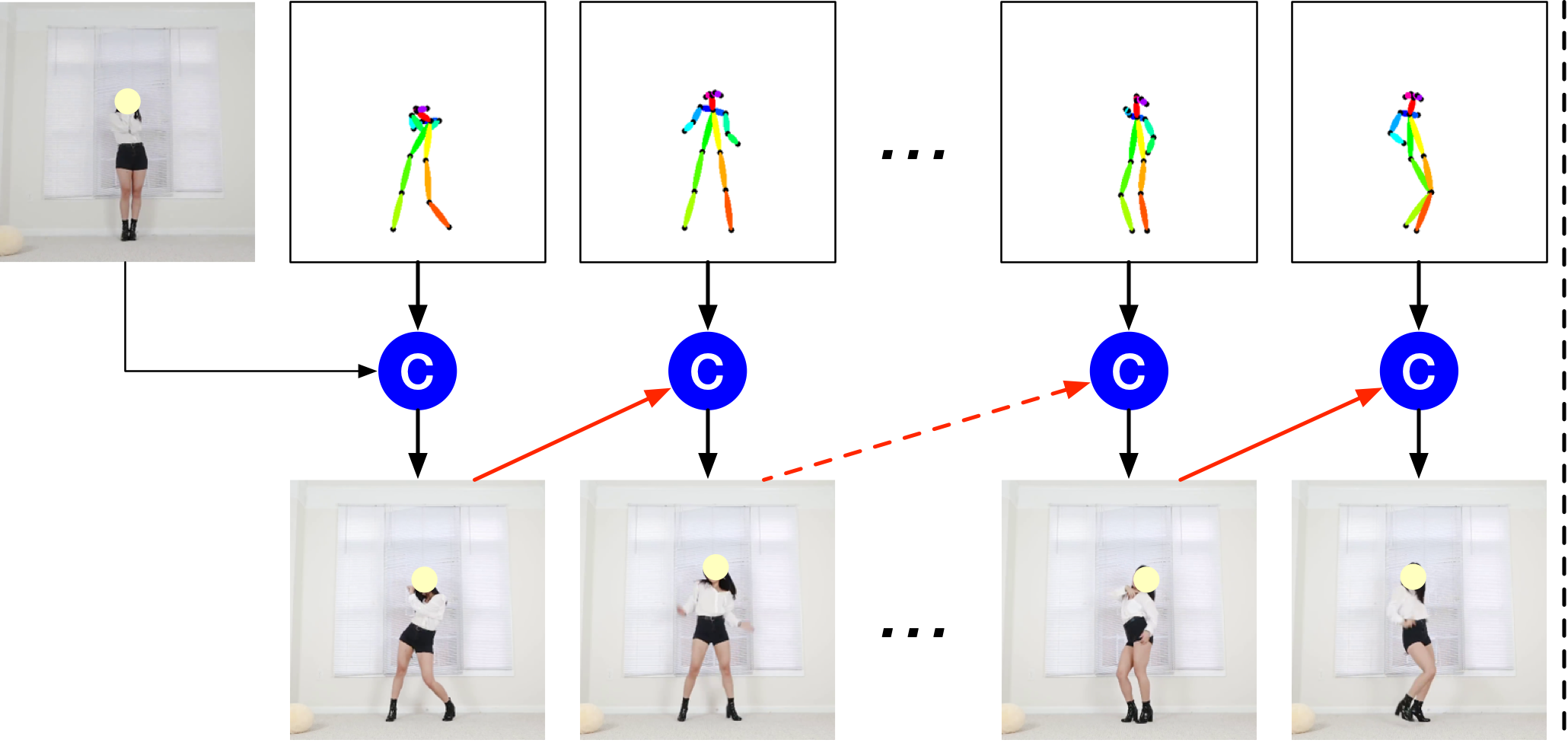}
	\includegraphics[width=0.32\textwidth]{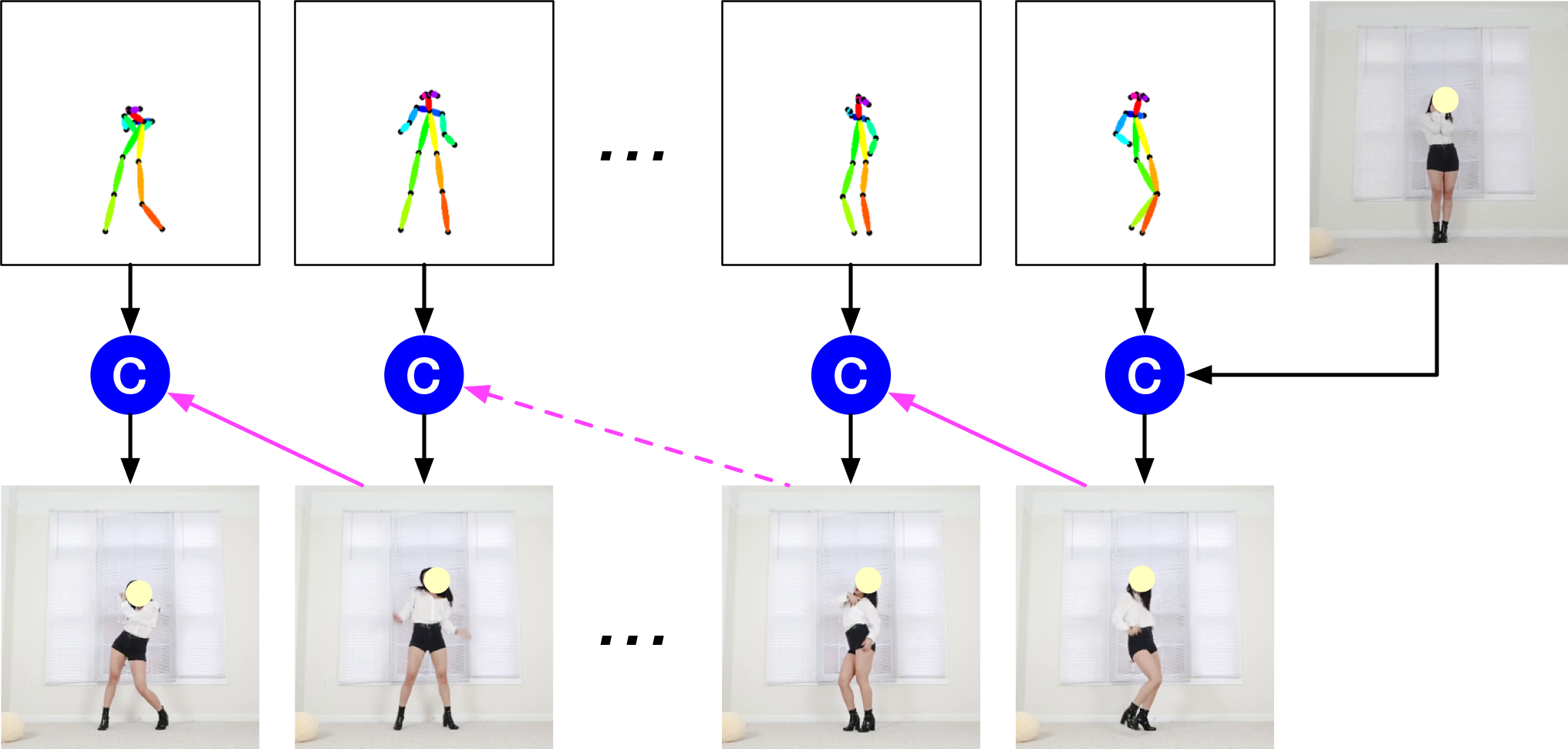}
	\caption{Different generation strategies for \hao{skeleton-to-video} translation.}
	\label{fig:stage2}
\end{figure*}

\noindent \textbf{Music Feature Extraction Network.}
We first divide the input music $M$ into 0.1s pieces. Next, these pieces are fed into the music feature extraction network $E$ to extract the audio features.
The network $E$ consists of an audio encoder and a two-layer bi-directional GRU \cite{cho2014properties}.
The output of $E$ is the hidden token $H_0 {\in} \mathbb{R}^{F{\times} h}$, where $h$ denotes depth.
Next, the hidden token $H_0$ is fed to a series of STGM blocks to produce a skeleton sequence $S$.

\hao{Although our Mamba-based method is used to process skeleton sequences, we chose to employ a GRU for the audio encoder because GRUs have been proven to be effective in capturing sequential patterns in audio signals with relatively low computational overhead. Our experiments suggest that a GRU is sufficient for modeling the audio context, while the main challenge lies in translating this audio into detailed dance motions—a task that benefits greatly from the capabilities of our Mamba-based model. Future work may explore integrating state-space or Transformer-based modules into the audio encoder, but for our current approach, the GRU offers a good balance between performance and complexity.}

\noindent \textbf{Spatial-Temporal Graph Mamba Block.}
After acquiring the hidden token $H_0$, we channel it through a succession of spatial-temporal graph Mamba (STGM) blocks. Within each STGM block, we integrate layer normalization (LN), graph convolution, state space modeling (SSM), and residual connections. The architecture of our proposed STGM block is illustrated in Figure~\ref{fig:block} (left).
To further elucidate, the LN ensures stable training dynamics by normalizing the inputs to each layer, mitigating the vanishing or exploding gradient issues commonly encountered in deep neural networks. Graph convolution enables effective information exchange between nodes in the spatial-temporal graph, facilitating the incorporation of context-aware features into the model. The integration of state space modeling introduces dynamic spatial-temporal relationships into the graph, enhancing the model's capability to capture complex spatial-temporal dependencies. 

Therefore, the output of the STGM blcok can be expressed as follows:
\begin{equation}
\begin{aligned}
    Z'_l & = GraphConv1d(MLP(LN(Z_{l-1}))), \\
    Z_l  & = MLP(LN(SG\_SSM(\sigma(Z'_l))) \times \sigma(MLP(LN(Z_{l-1}))) \\
    &+   LN(TGF\_SSM(\sigma(Z'_l))) \times \sigma(MLP(LN(Z_{l-1}))) \\
    &+ LN(TGB\_SSM(\sigma(Z'_l))) \times \sigma(MLP(LN(Z_{l-1})))),
\end{aligned}
\end{equation}
where $Z_l$ is the output of the $l$-th block and $Z_0$ is equal to $H_0$.
The $SG\_SSM(\cdot)$, $TGF\_SSM(\cdot)$, and $TGB\_SSM(\cdot)$ are spatial graph SSM, temporal graph forward SSM, and temporal graph backward SSM, respectively, they are the core of the proposed STGM block.

In Figure~\ref{fig:block} (right), we show how these three SSMs capture spatial and temporal relationships.
For the SG-SSM, we only present the $f$-th (where $f {=} 0, {\cdots}, F$) frame in Figure~\ref{fig:block} (top right).
The selection mechanism in SSM has a significant impact on spatial graphs by influencing the incorporation of relevant features and variables into the model. By selectively choosing which components of the spatial graph are informative for estimating the latent states of the system, the selection mechanism enhances the model's ability to capture spatial dependencies and dynamics effectively. This mechanism involves identifying the key nodes, edges, or graph structures that contribute to the evolution of the underlying state dynamics. Various techniques, such as feature selection algorithms or graph pruning methods, can be employed to facilitate this process. Through careful selection, the model can focus on the most relevant spatial information, leading to improved accuracy and efficiency in capturing the spatial relationships within the data. Additionally, the selection mechanism helps to reduce noise and redundancy within the spatial graph, resulting in more concise and informative representations of the underlying spatial processes.
Let $n_f^i$ be the $i$-th (where $i {=} 0, {\cdots}, V$) node in the $f$-th frame.
The goal of SG-SSM is to model the spatial correlations between joints/nodes within each frame using the selection mechanism, obtaining a new node position $\hat{n}_f^i$ with the global view of the spatial structure.
SG-SSM can selectively adjust themselves according to the spatial correlation, improving the representations of human structure and semantic consistency.

To generate a coherent skeleton sequence, the same nodes in each frame should be associated with each other. 
We build the TGF-SSM and TGB-SSM to explicitly and bidirectionally model the dependencies and learn the correlations across different frames.
The structures of TGF-SSM and TGB-SSM are illustrated in Figure~\ref{fig:block} (right middle and bottom). 
Inspired by the principles of Mamba, our temporal graph SSM architecture is meticulously crafted to traverse the input sequence in dual directions, thereby enhancing the model's resilience to permutation challenges. This capability becomes particularly salient in scenarios where implicitly ordered tokenization on graphs is not employed, ensuring that the model remains adept at extracting meaningful patterns regardless of the input's arrangement.
This bidirectional scanning mechanism imbues our architecture with a nuanced understanding of temporal relationships, enabling it to discern subtle patterns and dependencies that may exist within the data. By leveraging this dual-directional approach, our model exhibits a heightened ability to capture the intricate dynamics present in sequential data, bolstering its robustness and efficacy across a diverse range of tasks and datasets.

\noindent \textbf{Generation Head.}
We then apply a concatenation operation and convolution operation on the output to obtain the final sequence $S$.

\noindent \textbf{Optimization Objective.}
We use three losses as our full optimization objective,
$
\mathcal{L}_1 = \lambda_{p} \mathcal{L}_{p} + \lambda_{f} \mathcal{L}_{f} + \lambda_{l1} \mathcal{L}_{l1}
$,
where $\mathcal{L}_{p}$, $\mathcal{L}_{f}$, and $\mathcal{L}_{l1}$ denote the pose perceptual loss \cite{ren2020self}, feature matching loss \cite{wang2018high}, and $L_1$ reconstruction loss, respectively.

\subsection{\hao{Skeleton-to-Video} Translation}
\noindent \textbf{Overview.}
To translate the generated skeleton sequence~$S$ to a realistic dance video, we propose a new self-supervised learning regularization network $G_r$, as shown in Figure~\ref{fig:method}.
We use Vid2Vid \cite{wang2018video} as the backbone network of $G_r$.
Precisely, we follow the exact same implementation details (e.g., learning rate and training epochs) as Vid2Vid, but add our self-supervised regularization network $G_r$ on top to further improve \hao{skeleton-to-video} translation performance.
$G_r$ takes $S$ and a conditional image $I_0$ as input and aims to output video frames $I_i$ (where $i{=}1,2,{\cdots},F$) using three different generation strategies, as shown in Figure~\ref{fig:stage2}.
In this way, the conditional image $I_0$ can provide appearance content, while the skeleton sequence $S$ can provide motion information to simultaneously generate the final video frames~$I_i$.
Moreover, our model can generate different subject dance videos with the same music as input.

\noindent \textbf{Baseline Generation Strategy.}
In Figure~\ref{fig:stage2} (left), we combine each skeleton and the conditional image as the input to the network to generate the corresponding images. This baseline generation strategy can be formulated as follows,
\begin{equation}
\begin{aligned}
I_i = G_r({\rm concat}(I_0, S_i)), \quad i{=}1,2,{\cdots},F.
\end{aligned}
\end{equation}
However, this basic generation strategy cannot produce realistic images.
Thus, we introduce two new self-supervised regularizations to boost the quality of the generated images, i.e., forward/backward self-supervised regularizations (FSR/BSR).

\noindent \textbf{Forward Self-Supervised Regularization.}
We first generate images by using the forward generation strategy (Figure~\ref{fig:stage2} (middle)).
Specifically, we use the previous adjacent frame to generate the next frame, since the difference between the two adjacent frames is smaller and easier to learn. This generation strategy can be represented as follows,
\begin{equation}
\begin{aligned}
\hat{I}_i = G_r ({\rm concat}( I_{i-1}, S_i)), \quad i{=}2,
{\cdots},F,
\end{aligned}
\end{equation}
where $I_{i-1} {=} G_r ({\rm concat}(I_0, S_{i-1}))$. We further propose the FSR to reduce the difference between $I_i$ and $\hat{I}_i$; this can be formulated as follows,
\begin{equation}
\begin{aligned}
\mathcal{L}_{fsr} = ||I_i - \hat{I}_i||_1, \quad i{=}2,{\cdots},F.
\end{aligned}
\label{eq:fsr}
\end{equation}
By doing so, more constraints can be added to the network $G_r$ to generate more realistic and coherent video frames.

\noindent \textbf{Backward Self-Supervised Regularization.}
Similar to FSR, we propose BSR to further improve the feature representation of $G_r$ in the backward direction.
Specifically, we first generate images using the backward generation strategy (Figure~\ref{fig:stage2} (right)),
\begin{equation}
\begin{aligned}
\tilde{I}_i = G_r ({\rm concat}( I_{i+1}, S_i)), \quad i{=}1, 2,{\cdots},F{-}1,
\end{aligned}
\end{equation}
where $I_{i+1} {=} G_r ({\rm concat}(I_0, S_{i+1}))$. We further propose the BSR to reduce the difference between $I_i$ and $\tilde{I}_i$,
\begin{equation}
\begin{aligned}
\mathcal{L}_{bsr} = ||I_i - \tilde{I}_i||_1, \quad i{=}1,2,{\cdots},F{-}1.
\end{aligned}
\label{eq:bsr}
\end{equation}
Note that BSR and FSR are different. They are proposed to ensure that the forward and backward generation results are consistent. We see in Table \ref{tab:abla2} that when we use FSR and BSR simultaneously, the result is much better than using each one alone, which means that both FSR and BSR are necessary.
Also, BSR and FSR are proposed to reduce error accumulation. The motivation behind this is to make the video result of the forward generation consistent with that of the backward generation, which needs the model to continuously reduce the accumulation of errors.

We use previous/subsequent frames as forward/backward constraints during the training.
During testing, we only use the baseline generation strategy to produce the final results since FSR and BSR can be regarded as two losses to constrain the model during training, and in testing, we only chose the basic generation strategy to get the results for efficiency.

\noindent \textbf{Optimization Objective.}
The optimization objective can be written as follows,
$
\mathcal{L}_2 = \lambda_{gan} \mathcal{L}_{gan} + \lambda_{ l1} (\mathcal{L}_{l1} + \mathcal{L}_{fsr} + \mathcal{L}_{bsr})
$,
where $\mathcal{L}_{l1}{=}||I_i -I^{real}_i||_1$ ($i{=}1,2,{\cdots},F$) and $I^{real}_i$ are the real images. $\mathcal{L}_{gan}$ is the adversarial loss between the generated image $I_i$ and the real one $I^{real}_i$. We set $\lambda_{gan}{=}1$ and $\lambda_{l1}{=}10$ in our experiments.

\section{Experiments}

\begin{figure}[!t] \small
	\centering
\includegraphics[width=1\linewidth,height=0.04\linewidth]{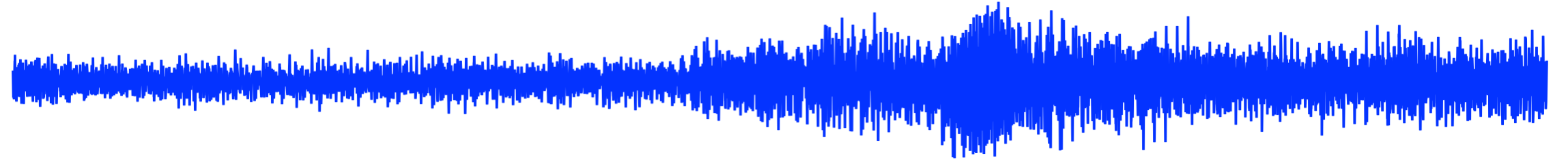}
	\includegraphics[width=1\linewidth]{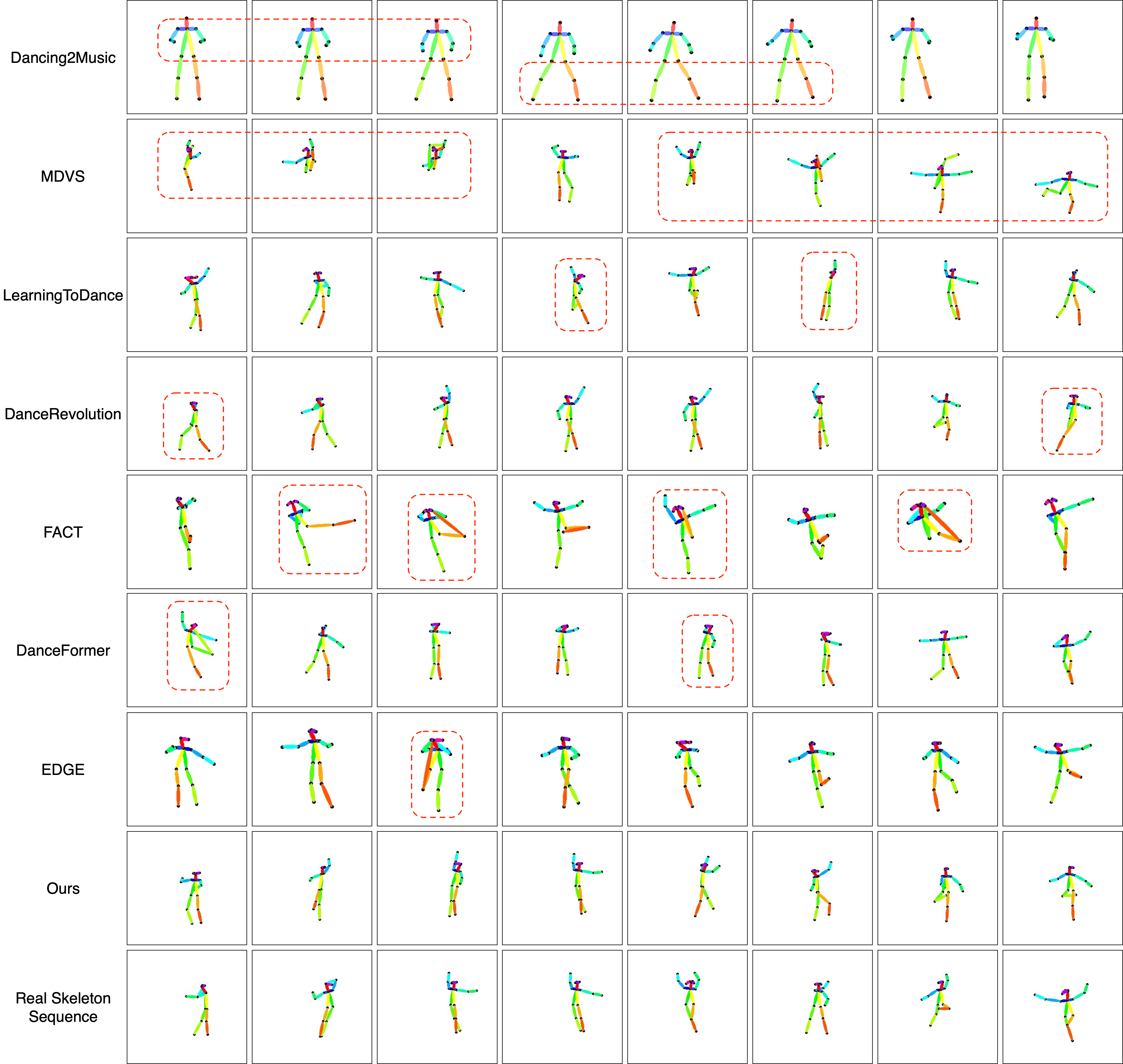}
	\caption{\hao{Visual comparisons for music-to-skeleton translation.  Red dashed boxes above two frames indicate that the skeletons are repetitive. If the red dashed box is only on a certain frame, it means that the skeleton is unnatural or distorted.
	}}
	\label{fig:sota1}
\end{figure}

\begin{figure}[!ht] \small
	\centering
\includegraphics[width=1\linewidth,height=0.05\linewidth]{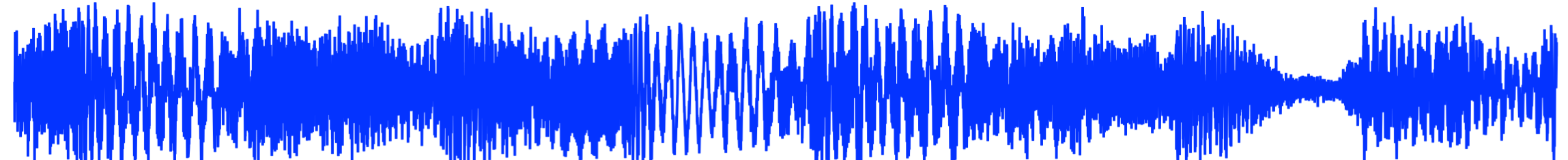}
	\includegraphics[width=1\linewidth]{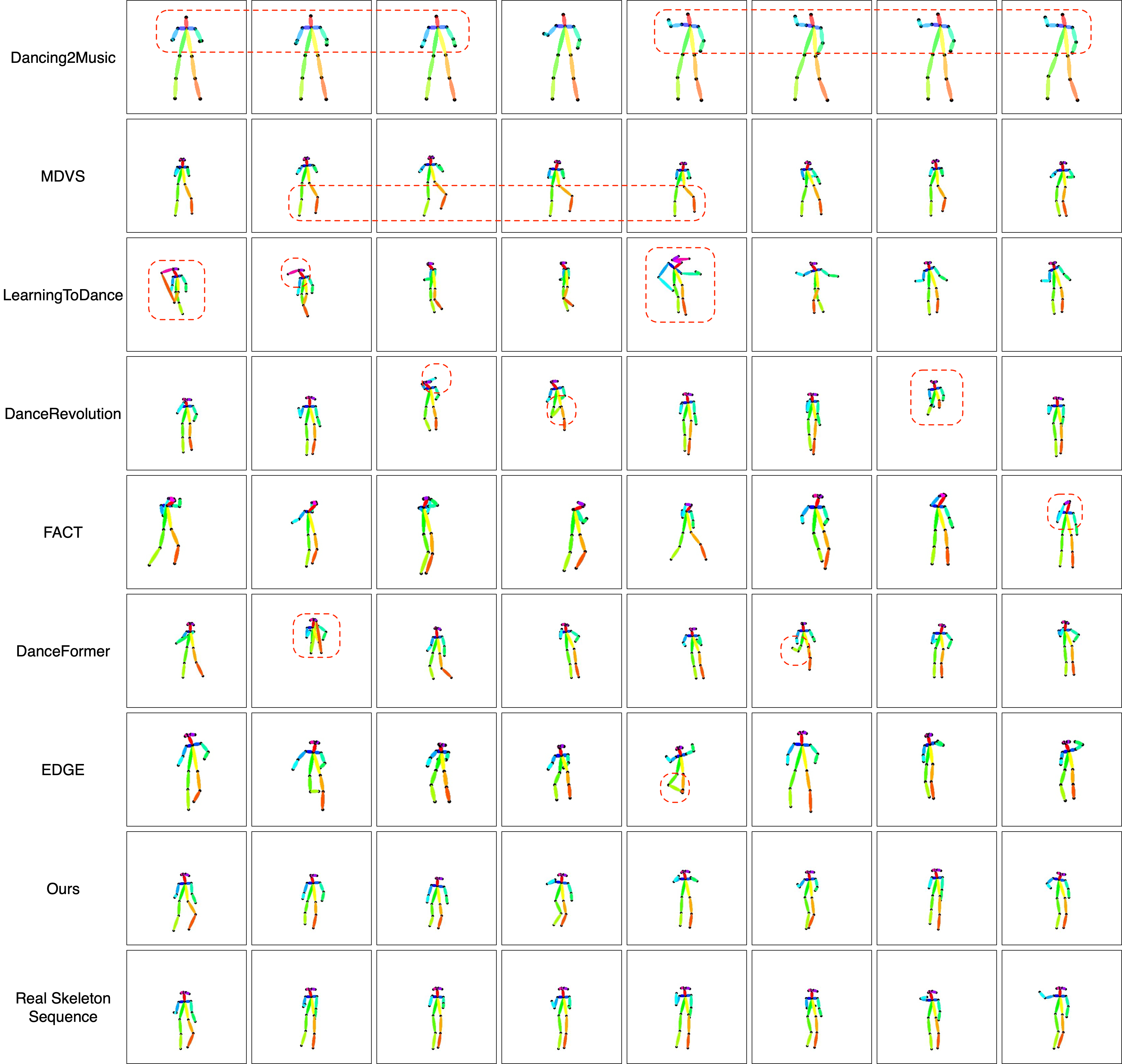}
	\caption{\hao{Visual comparisons for music-to-skeleton translation.  Red dashed boxes above two frames indicate that the skeletons are repetitive. If the red dashed box is only on a certain frame, it means that the skeleton is unnatural or distorted.
	}}
	\label{fig:supp1}
\end{figure}

\subsection{Music-to-Skeleton Translation}
\noindent \textbf{Datasets.}
We use the dataset proposed in \cite{ren2020self} for a fair comparison, which contains 100 videos, i.e., 40 k-pop videos, 20 ballet videos, and 40 popping videos. 
We then use OpenPose \cite{cao2017realtime} to extract human skeletons, resulting in 1,782 k-pop clips, 448 ballet clips, and 1518 popping clips in total.
We follow \cite{ren2020self} and select the last 10\% of each type of dance for testing, and the remaining videos are used for training.
We train our model for 100 epochs with a batch size of 16.

\begin{figure}[!t] \small
	\centering
\includegraphics[width=1\linewidth,height=0.05\linewidth]{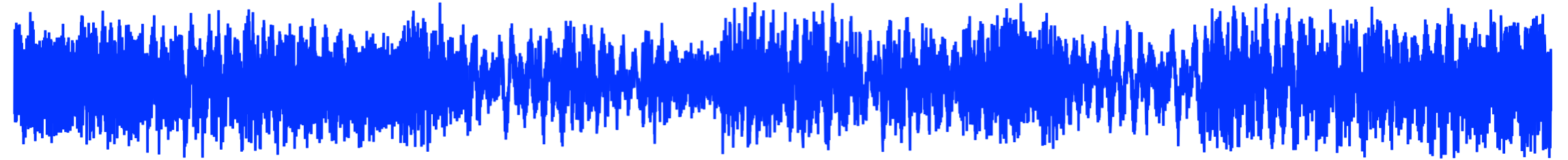}
	\includegraphics[width=1\linewidth]{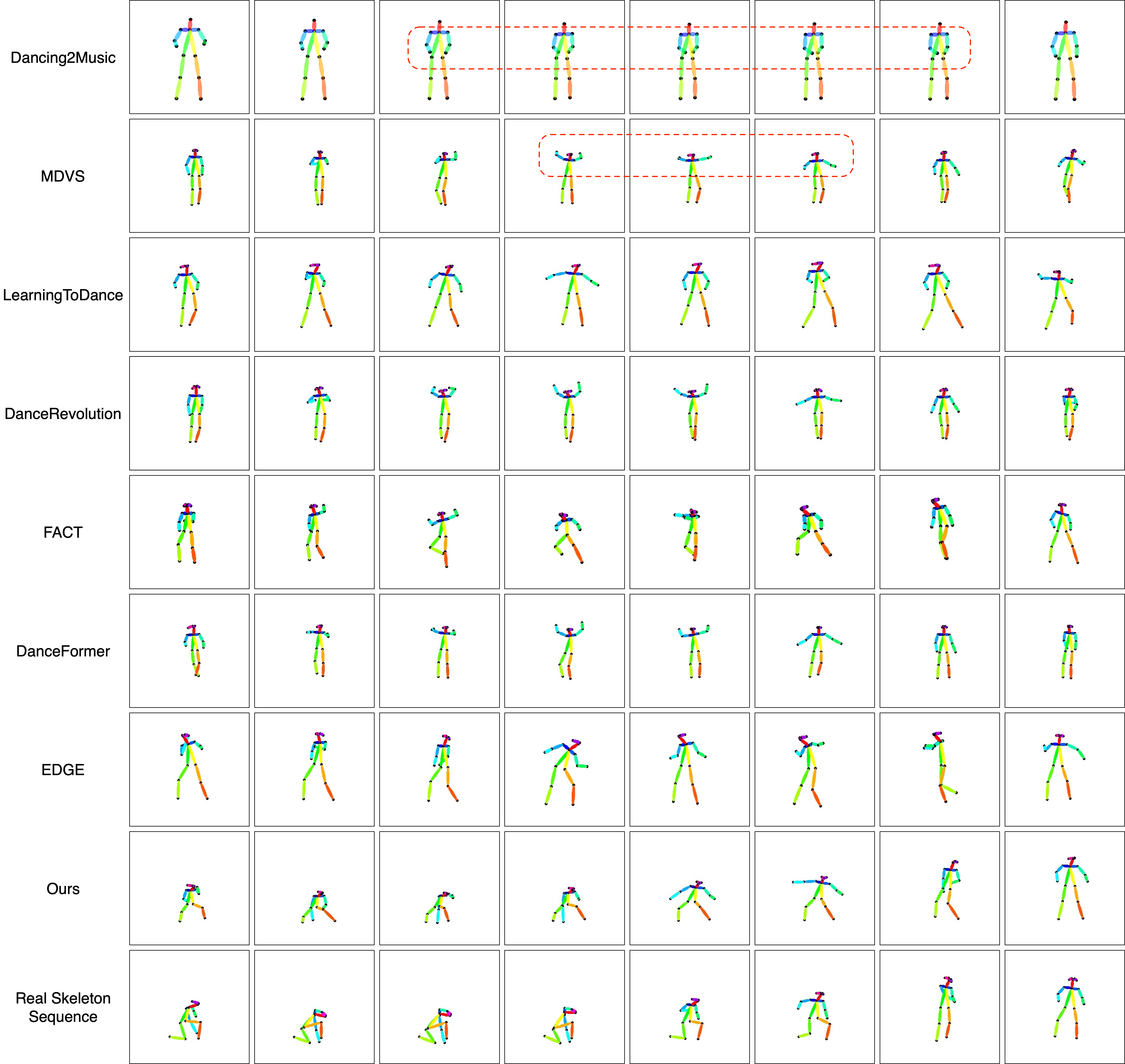}
	\caption{\hao{Visual comparisons for music-to-skeleton translation.  Red dashed boxes above two frames indicate that the skeletons are repetitive. If the red dashed box is only on a certain frame, it means that the skeleton is unnatural or distorted.
	}}
	\label{fig:supp2}
\end{figure}

\noindent \textbf{Qualitative Evaluation.}
We compare our method with several leading music-to-skeleton methods, i.e., Dancing2Music \cite{lee2019dancing}, MDVS \cite{ren2020self}, LearningToDance \cite{ferreira2021learning}, DanceRevolution \cite{huang2020dance}, DanceFormer \cite{li2022danceformer}, \hao{FACT \cite{li2021ai}, and EDGE \cite{tseng2023edge}}.
We used the pre-trained models provided by the authors to generate the results. 
For methods for which the authors did not provide source code, we try our best to reimplement those models for a fair comparison.
The results are shown in Figures~\ref{fig:sota1}, \ref{fig:supp1}, and \ref{fig:supp2}. It is easy to see that those existing methods generate jerking dances that are prone to repeating the same movements (as shown in the red dashed boxes).
Compared to existing methods, our results are more realistic and coherent.

\noindent \textbf{User Study.}
To evaluate the quality of the generated skeleton sequences, we also conduct a user study via Amazon Mechanical Turk.
Specifically, users are first asked to answer a background question: ``Have you learned to dance or have experience dancing?''. Based on their answers, they are labeled as ``Expert'' or ``Non-Expert''. 
Then, given 8 five-second skeleton sequences (6 generated by existing methods, 1 generated by our proposed method, and 1 real skeleton sequence) with input music, 30 participants including 12 dancers need to answer two questions: ``Which skeleton sequence is more realistic regardless of the music?'' and ``Which skeleton sequence matches the music better?''. Users have unlimited time to select their choices for 40 trials.
For fair evaluation, we inform users in advance that there are two forms of skeleton, since Dancing2Music \cite{lee2019dancing} produces a slightly different form from the others.
The results of our model compared to existing methods and the real skeleton sequence are shown in Table~\ref{tab:amt1}.
Users show more preference for our approach in terms of both motion realism and style consistency than the other methods, indicating the effectiveness of the proposed method.

\begin{table}[!t] \small
	\centering
	\caption{User study (motion realism (\%) and style consistency  (\%)), beat coverage  (\%), and beat hit rate  (\%) results for music-to-skeleton translation.
	}
	\resizebox{1\linewidth}{!}{%
		\begin{tabular}{rcccccc} \toprule
			\multirow{2}{*}{Method}  & \multicolumn{2}{c}{Motion Realism $\uparrow$} & \multicolumn{2}{c}{Style Consistency $\uparrow$} &  \multirow{2}{*}{Beat Coverage $\uparrow$} & \multirow{2}{*}{Beat Hit Rate $\uparrow$} \\ \cmidrule(lr){2-3} \cmidrule(lr){4-5}
			& Expert    & Non-Expert & Expert & Non-Expert  \\ \midrule
		Dancing2Music \cite{lee2019dancing}          & 1.2 & 1.6 & 2.1 & 2.3 & 16.7 & 32.1 \\
		MDVS \cite{ren2020self}                      & 2.7 & 3.2 & 3.2 & 4.1 & 23.5 & 48.8 \\
		LearningToDance \cite{ferreira2021learning}  & 5.2 & 6.6& 5.8& 6.5 &  33.7 & 52.4\\
		DanceRevolution \cite{huang2020dance}  & 8.1  & 8.5& 7.9 & 7.9& 37.6 & 59.6 \\
            \hao{FACT \cite{li2021ai}} & \hao{11.7} & \hao{11.8} & \hao{10.4} & \hao{9.3} & \hao{38.2} & \hao{63.7} \\
		DanceFormer \cite{li2022danceformer}    &13.3 &  13.9 & 13.4& 12.4& 40.9 & 68.5 \\
            \hao{EDGE \cite{tseng2023edge}} & \hao{16.1} & \hao{15.8} & \hao{15.7} & \hao{15.8} & \hao{48.3} & \hao{74.8} \\
		STG-Mamba (Ours)          & \textbf{19.6} & \textbf{18.4} & \textbf{19.5} & \textbf{19.2} & \textbf{58.1} & \textbf{81.9} \\ \hline
		Real Sequence             & 22.1 & 20.2 & 22.0 & 22.5  & - & - \\ \bottomrule
	\end{tabular}}
	\label{tab:amt1}
\end{table}

\begin{table}[!t]
	\centering
	\caption{PFD, VFD, PVar, and BC results for music-to-skeleton translation. Our method surpasses other baselines in terms of quality, diversity, and beat consistency.}
	\resizebox{1\linewidth}{!}{%
		\begin{tabular}{rccccc} \toprule
			Method  & PFD $\downarrow$ & VFD $\uparrow$ & PVar $\uparrow$ & BC $\uparrow$ & \hao{Inference Time (s) $\downarrow$} \\ \midrule 
			Dancing2Music \cite{lee2019dancing}  &932.1&0.93&0.148& 0.194 \\
			MDVS \cite{ren2020self}     &653.8&2.20&0.259& 0.246 & \hao{-} \\
			LearningToDance \cite{ferreira2021learning} &426.6& 2.86&0.364& 0.313 & \hao{-} \\
		DanceRevolution \cite{huang2020dance}    &374.5&3.51& 0.406&  0.354 & \hao{-}\\
            \hao{FACT \cite{li2021ai}} & \hao{321.4} & \hao{3.78} & \hao{0.434} & \hao{0.382} & \hao{3.8}\\
			DanceFormer \cite{li2022danceformer}        & 225.7 & 3.94 & 0.542 & 0.451 & \hao{4.2} \\
            \hao{EDGE \cite{tseng2023edge}} & \hao{145.9} & \hao{4.37} & \hao{0.681} & \hao{0.547} & \hao{4.8}\\
			STG-Mamba (Ours) & \textbf{86.2} & \textbf{4.82} & \textbf{0.875} & \textbf{0.684} & \hao{2.2}\\ \bottomrule
	\end{tabular}}
	\label{tab:sota1}
\end{table}

\noindent \textbf{Quantitative Evaluation.}
We follow DanceFormer \cite{li2022danceformer} and employ Frechet Distance (PFD and VFD), diversity score (i.e., PVar), and bet consistency score (BC) as our evaluation metrics.
\hao{Our method based-on Mamba, which achieves O(N) complexity by imposing certain constraints. However, these constraints also help in enforcing a more structured representation of the sequence dynamics, which is beneficial for our task. The results compared to existing methods (including Transformer-based methods, e.g., FACT, DanceFormer, and EDGE) in Table~\ref{tab:sota1} show that this state-space formulation, designed to capture sequential dynamics in a more focused way, actually outperforms standard attention mechanisms when it comes to modeling nuanced movements in dance sequences.}

\noindent \textbf{\hao{Inference Time.}} 
\hao{
We also provide an inference time comparison between our method, FACT \cite{li2021ai}, DanceFormer \cite{li2022danceformer}, and EDGE \cite{tseng2023edge} in Table~\ref{tab:sota1}. As shown, our method has the shortest inference time among the compared methods, demonstrating its efficiency in inference speed. Notably, FACT, DanceFormer, and EDGE are all Transformer-based methods, which typically involve higher computational overhead due to their attention mechanisms. In contrast, our method leverages a Mamba-based architecture, which is designed to be more efficient by reducing the computational complexity associated with attention operations. This architectural difference contributes to our method’s superior inference speed, making it particularly well-suited for real-time applications where low latency is critical.}

\begin{figure}[!t] \small
	\centering
	\includegraphics[width=1\linewidth]{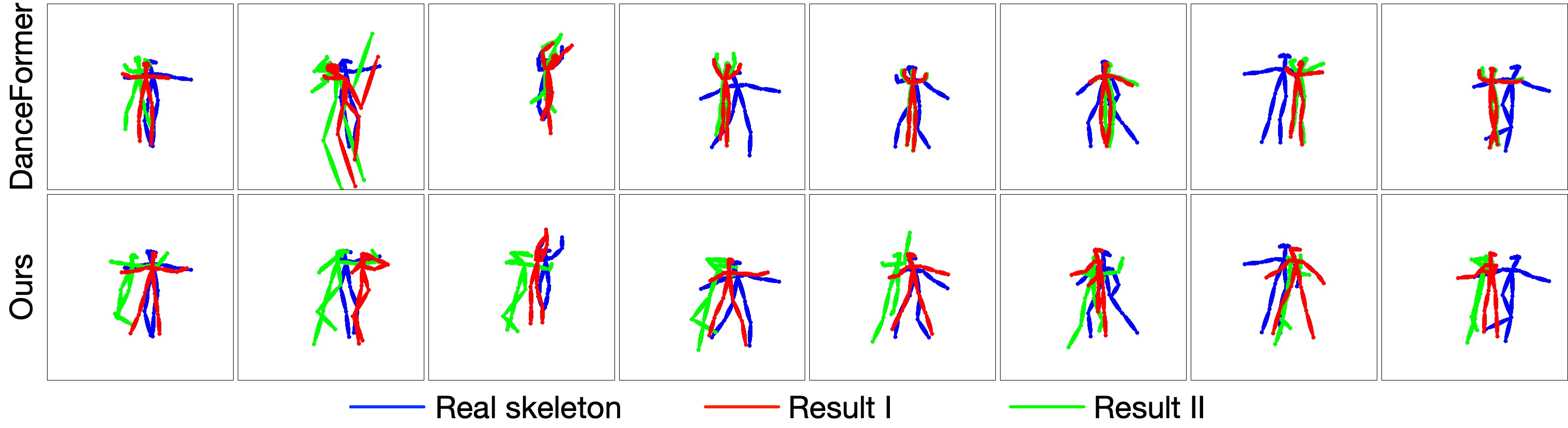}
	\caption{Our multi-modal results vs. DanceFormer. Zoom-in for the best view.}
	\label{fig:rebuttal1}
\end{figure}

\noindent \textbf{Multi-Modal Generation.} 
\hao{Multi-modal generation means that by inputting single music, we aim to get both realistic and diverse skeletons by adding some random noise.
We observe that the generated sequence and the real one are different but share the same beat and style, since our model generates basic movements that adjust to the rhythm.
In Figure~\ref{fig:rebuttal1}, we have overlaid skeletons to show how the sequences differ.
We see that our generated skeletons differ from the real ones. 
We also provide the comparison results with the SOTA method DanceFormer in Figure~\ref{fig:rebuttal1}. Our method generates more diverse results, while DanceFormer tends to generate repeated results.}

\begin{figure}[!t] \small
	\centering
	\includegraphics[width=1\linewidth]{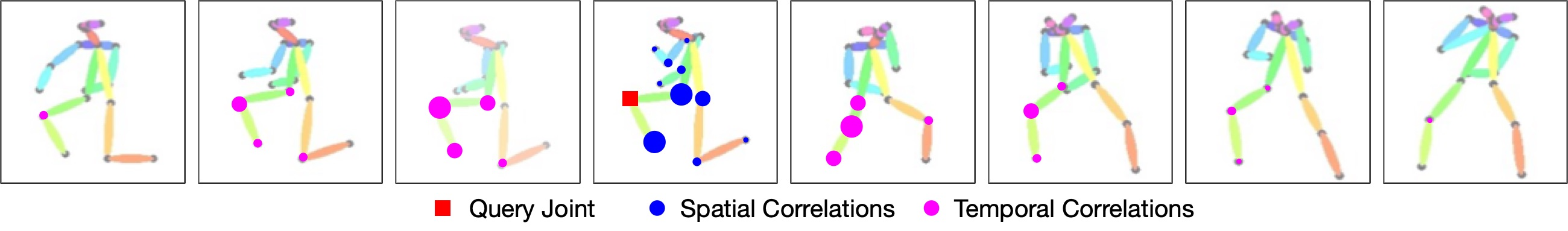}
	\caption{Visualization of the responses of spatial (top-10 joints) and temporal (top-20 joints) correlations. The circles' size indicate the response magnitudes.}
	\label{fig:corr}
\end{figure}

\noindent \textbf{Visualization of Learned Spatio-Temporal Dependences.}
We visualize the learned spatial and temporal dependences in Figure~\ref{fig:corr}, where the circle around each joint indicates the magnitude of feature responses of this joint to the example joint. Our spatial and temporal graph SSMs indeed learn the relationship of joints spatially and temporally. 

\subsection{\hao{Skeleton-to-Video} Translation}

\begin{figure}[!t] \small
	\centering
	\includegraphics[width=0.8\linewidth]{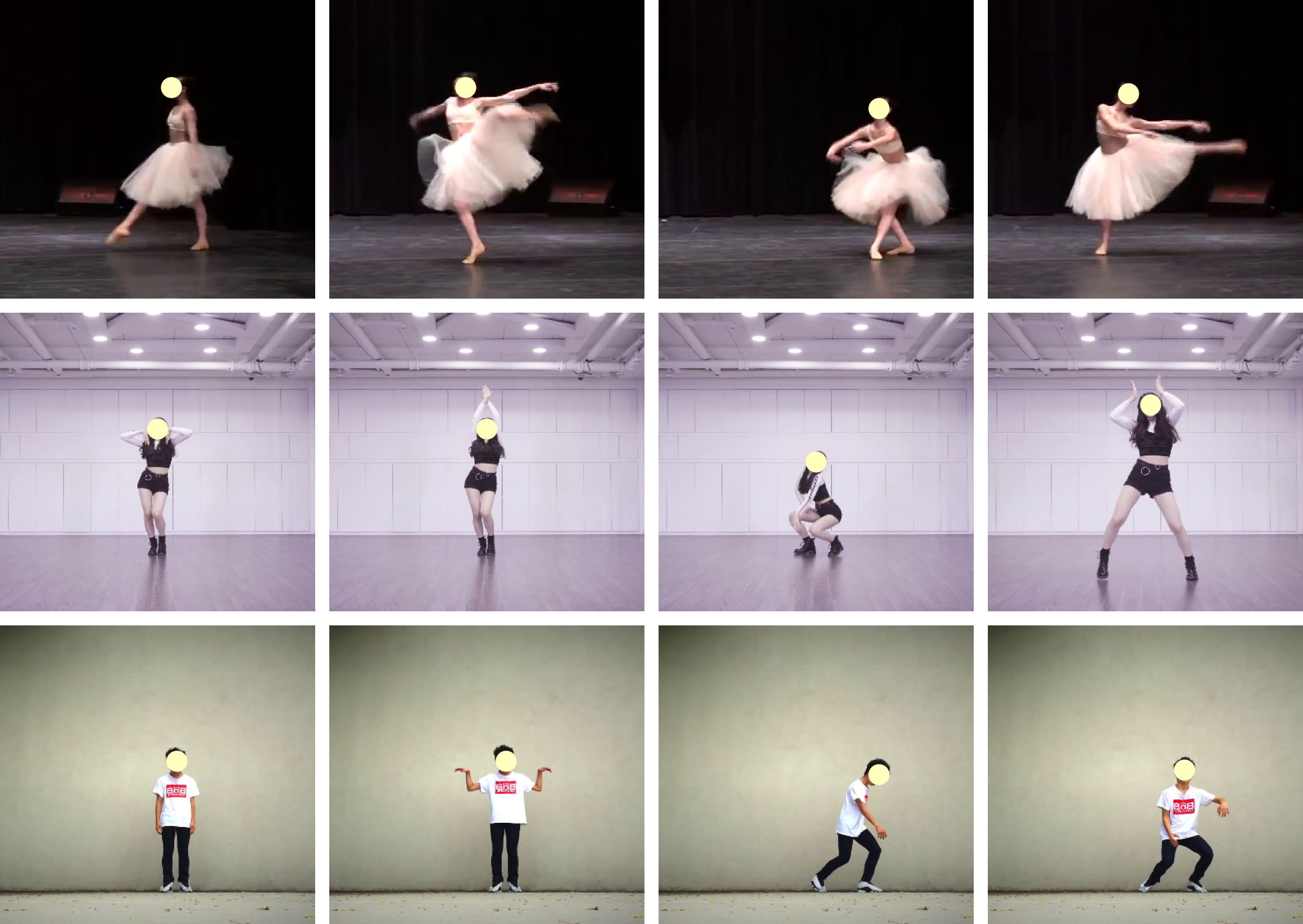}
	\caption{Examples of video frames collected from the internet for the \hao{skeleton-to-video} translation task.}
	\label{fig:dataset}
\end{figure}

\noindent \textbf{Datasets.}
The authors of \cite{ren2020self} cannot release the training images due to data privacy.
Therefore, to train the proposed model, we collected training data from Internet.
We have three types of dances (ballet, popping, and kpop), with 62\% female and 38\% male dancers and 76\% indoor and 24\% outdoor scenes. Figure~\ref{fig:dataset} shows several examples of the collected dataset.
We employ OpenPose \cite{cao2017realtime} as our pose joints detector and filter out images where no human bodies were detected, as well as those smaller than $256{\times}256$. We then resize the images to $256{\times}256$.
In total, we obtain 54,944 video sequences, each of which contains the extracted skeletons and the corresponding images. 
\hao{These 54,944 video sequences are extracted from a limited number of original videos. Each sequence is a continuous segment, preserving motion continuity. When splitting the dataset into training and test sets, we ensured that no overlapping segments from the same original video appear in both sets. This prevents data leakage and ensures a fair evaluation of the model’s generalization ability. In total, 4,581 sequences are used for testing, while the remaining 50,363 sequences are used for training.}
We train our model for 40 epochs with a batch size of 8.

\begin{figure}[tbp] \small
	\centering
	\includegraphics[width=1\linewidth]{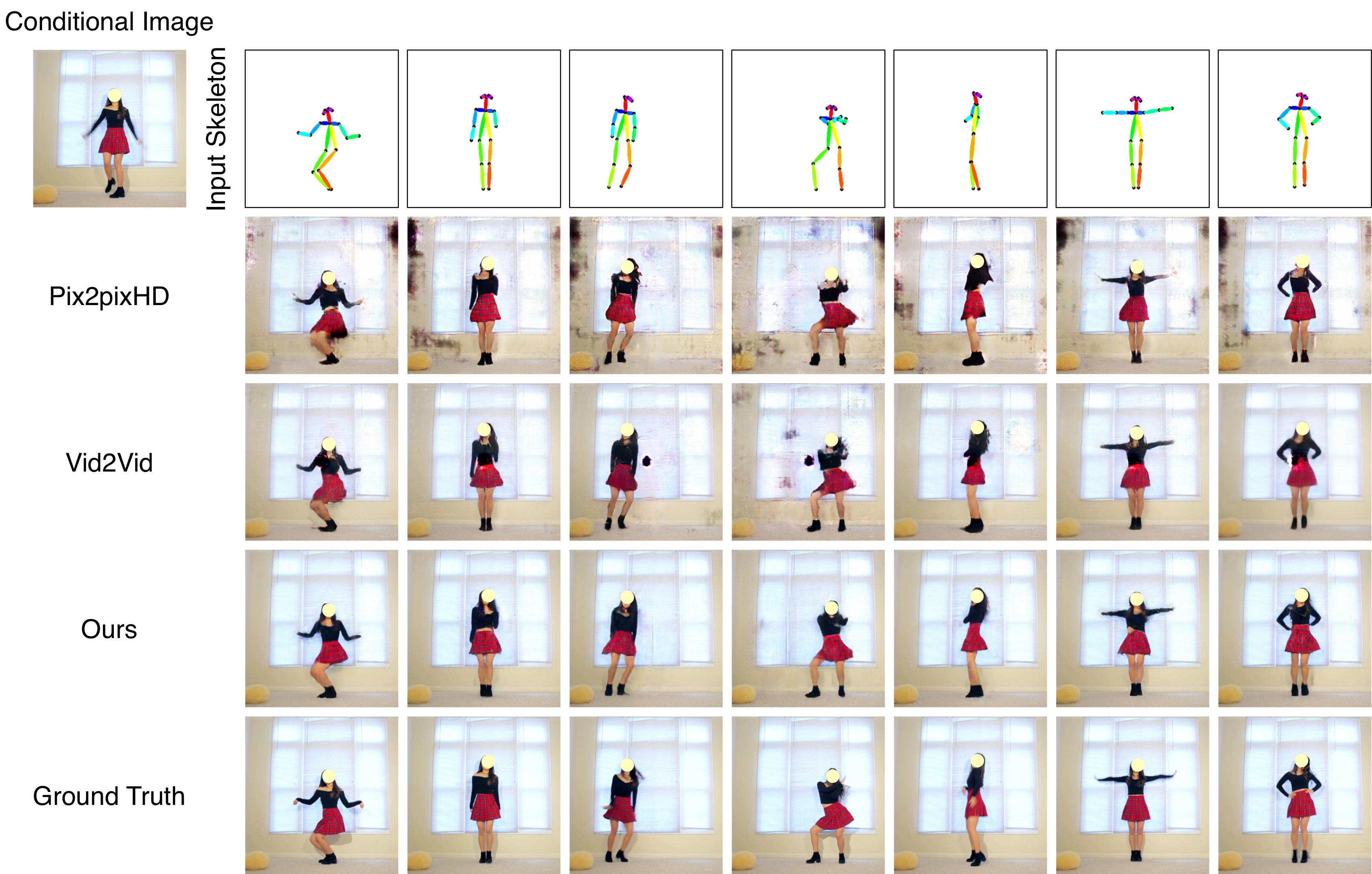}
	\caption{Visual comparisons for \hao{skeleton-to-video} translation.}
	\label{fig:sota2}
\end{figure}

\noindent \textbf{Qualitative Evaluation.}
Existing methods such as Dancing2Music \cite{lee2019dancing}, and MDVS \cite{ren2020self} directly employ existing motion transfer models to convert the generated skeleton sequence to realistic video frames.
Specifically, Dancing2Music \cite{lee2019dancing} adopts Vid2Vid \cite{wang2018video} to achieve this goal, while MDVS \cite{ren2020self} uses Pix2pixHD \cite{wang2018high} to translate skeleton sequences into video frames.
We compare our method with both leading methods in Figure~\ref{fig:sota2}. Our method generates more realistic and coherent video frames, further validating the effectiveness of the proposed self-supervised regularization network.

\noindent \textbf{Generation on Different Subjects.}
We also show the generation results conditioned on different subjects in Figure~\ref{fig:sota2_10}.
Specifically, given the input music, our proposed method first generates a skeleton sequence. 
Then, different conditional images are given to generate realistic dancing videos with the same style but different subjects. 

\noindent \textbf{User Study.}
We also conduct a user study to evaluate the generated images. Participants are asked to watch a series of video triplets with the real skeleton sequence (two videos are synthesized using Pix2pixHD and Vid2Vid, and the other by our method). 
They are then asked to pick the most realistic one and we give them unlimited time to respond. Each task consists of 20 video triplets and is performed by 30 distinct participants. 
The results are shown in Table~\ref{tab:sota2}.
More participants consider our results to be more realistic than those provided by the other methods.

\begin{table}[t]
	\centering
	\caption{User study, FID, LPIPS, and FVD results for \hao{skeleton-to-video} translation.}
	\resizebox{1\linewidth}{!}{%
		\begin{tabular}{rcccc} \toprule
			Method  & User Preference (\%) $\uparrow$ & FID $\downarrow$  & LPIPS $\downarrow$  & FVD $\downarrow$ \\ \midrule	
			Pix2pixHD \cite{wang2018high} & 28.9 & 61.68 & 0.2135 & 129.8 \\
			Vid2Vid \cite{wang2018video} & 30.6 & 47.65 & 0.1812   & 81.5 \\ 
			STG-Mamba (Ours)                         & \textbf{40.5} & \textbf{35.17} & \textbf{0.1529} & \textbf{45.4} \\ \bottomrule
	\end{tabular}}
	\label{tab:sota2}
\end{table}

\begin{table}[t]
	\centering
	\caption{Ablation study of music-to-skeleton translation.}
	\resizebox{1\linewidth}{!}{%
		\begin{tabular}{clcccc} \toprule
			\# & Setting & PFD $\downarrow$ & VFD $\uparrow$ & PVar $\uparrow$ & BC $\uparrow$  \\ \midrule	
			S1-1 & Baseline                 &653.8&2.20&0.259& 0.246 \\
			S1-2 & S1-1 + SG-SSM            & 428.3 &2.86 & 0.362 & 0.332\\
			S1-3 & S1-1 + TGF-SSM           & 326.9 & 3.52 & 0.523 & 0.513\\
			S1-4 & S1-1 + TGB-SSM           & 318.6 & 3.63 & 0.540 & 0.535\\ 
			S1-5 & S1-1 + TGF-SSM + TGB-SSM  & 130.7 & 4.47 & 0.761 & 0.629  \\ 
			S1-6 & S1-2 + TGF-SSM + TGB-SSM  &\textbf{86.2} & \textbf{4.82} & \textbf{0.875} & \textbf{0.684} \\
			\bottomrule
	\end{tabular}}
	\label{tab:abla1}
\end{table}

\begin{table}[t]
	\centering
	\caption{Ablation study of \hao{skeleton-to-video} translation.}
	\resizebox{0.8\linewidth}{!}{%
		\begin{tabular}{clccccc} \toprule
			\# & Setting  & FID $\downarrow$ & LPIPS $\downarrow$ &  FVD $\downarrow$ \\ \midrule	
			S2-1 & Baseline & 47.65  & 0.1812 & 81.5  \\
			S2-2 & S2-1 + FSR & 42.89 & 0.1785  & 57.3\\
			S2-3 & S2-1 + BSR & 40.15  & 0.1667  & 55.2\\
			S2-4 & S2-1 + FSR + BSR & \textbf{35.17} &\textbf{0.1529} & \textbf{45.4} \\  \bottomrule
	\end{tabular}}
	\label{tab:abla2}
\end{table}

\noindent \textbf{Quantitative Evaluation.}
We adopt FID \cite{heusel2017gans}, LPIPS \cite{zhang2018unreasonable}, and FVD \cite{unterthiner2019fvd} to evaluate \hao{skeleton-to-video} translation results.
The comparison results with two video generation methods, i.e., Vid2Vid \cite{wang2018video} and Pix2pixHD \cite{wang2018high}, are shown in Table~\ref{tab:sota2}. Our method achieves the best results in both evaluation metrics, confirming that the video frames generated by our method are more realistic.

\subsection{Ablation Study}
We conduct extensive ablation studies to evaluate the effectiveness of each component of the proposed method.

\noindent \textbf{Analysis for Music-to-Skeleton Translation.}
This translation process has six baselines, as shown in Table~\ref{tab:abla1}.
\hao{
We conducted an ablation study to replace the pose generator used in MDVS \cite{ren2020self} with our proposed SSM blocks. 
The details of our settings are as follows:
1) S1-1: This is our baseline, which uses the pose generator proposed in MDVS \cite{ren2020self}.
2) S1-2: We replace the pose generator with our proposed SG-SSM block to specifically model spatial correlations between different human joints.
3) S1-3: We substitute the pose generator with our TGF-SSM block, which captures long-range dependencies between joints in the forward temporal direction.
4) S1-4: We use the TGB-SSM block to improve joint correlations in the backward temporal dimension.
5) S1-5: We combine TGF-SSM and TGB-SSM to model temporal correlations from both directions.
6) S1-6: This is our full model, employing all three proposed SSM blocks (SG-SSM, TGF-SSM, and TGB-SSM). Compared to \cite{ren2020self}, which uses a self-supervised pose generator with a pose perceptual loss to generate skeleton sequences from music, our approach leverages these specialized SSM blocks to more effectively capture spatial and temporal dependencies. Our experiments demonstrate that each of these substitutions contributes to a more coherent and natural dance video synthesis.}

Table~\ref{tab:abla1} shows that SG-SSM, TGF-SSM and TGB-SSM improve the generation performance over the baseline, validating the effectiveness. 
In addition, by adding the proposed SSMs in the baseline S1-6, overall performance is further enhanced.

\noindent \textbf{Analysis of \hao{Skeleton-to-Video} Translation.}
The generation process has four baselines as shown in Table~\ref{tab:abla2}: 
1) S2-1 is our baseline (i.e., $\lambda_{l1} \mathcal{L}_{l1}$), aiming to generate video frames using the baseline generation strategy, i.e., Vid2Vid \cite{wang2018video}.
2) S2-2 adopts the proposed FSR to generate video frames using the proposed forward generation strategy, i.e., $\lambda_{l1} (\mathcal{L}_{l1} {+} \mathcal{L}_{fsr})$.
3) S2-3 generates video frames using the proposed backward generation strategy and the proposed BSR, i.e., $\lambda_{l1} (\mathcal{L}_{l1} {+} \mathcal{L}_{bsr})$.
4) S2-4 is our full model and employs both FSR and BSR to improve the performance of video generation in a self-supervised way, i.e., $\lambda_{l1} (\mathcal{L}_{l1} {+} \mathcal{L}_{fsr} {+} \mathcal{L}_{bsr})$.
Table~\ref{tab:abla2} confirms that FSR and BSR boost the generation performance over the baseline, validating the effectiveness of our regularization. 
Moreover, by introducing both FSR and BSR together in the baseline S2-4, the overall results are further improved.

\section{Conclusion}

Our proposed STG-Mamba method represents a significant advancement in the task of music-guided dance video synthesis. By introducing the novel STGM block and a self-supervised regularization network, we achieve a more effective construction of skeleton sequences from input music and their translation into high-quality dance videos. Furthermore, the collection of a large-scale dataset for \hao{skeleton-to-dance-video} translation provides valuable resources for future research endeavors. The experimental results demonstrate the superiority of STG-Mamba in various aspects, reaffirming the efficacy and practicality of our proposed approach. In the future, we plan to further explore the proposed STG-Mamba on other graph-related tasks.

\section*{Acknowledgements}
This work is partially supported by the Fundamental Research Funds for the Central Universities, Peking University. Nicu Sebe acknowledges funding by the European Union’s Horizon Europe research and innovation program under grant agreement No. 101120237 (ELIAS) as well as the support of the PNRR project FAIR - Future AI Research (PE00000013), under the NRRP MUR program funded by the NextGenerationEU.

\small
\bibliographystyle{IEEEtran}
\bibliography{ref}

\begin{IEEEbiography}[{\includegraphics[width=1in,height=1.25in,clip,keepaspectratio]{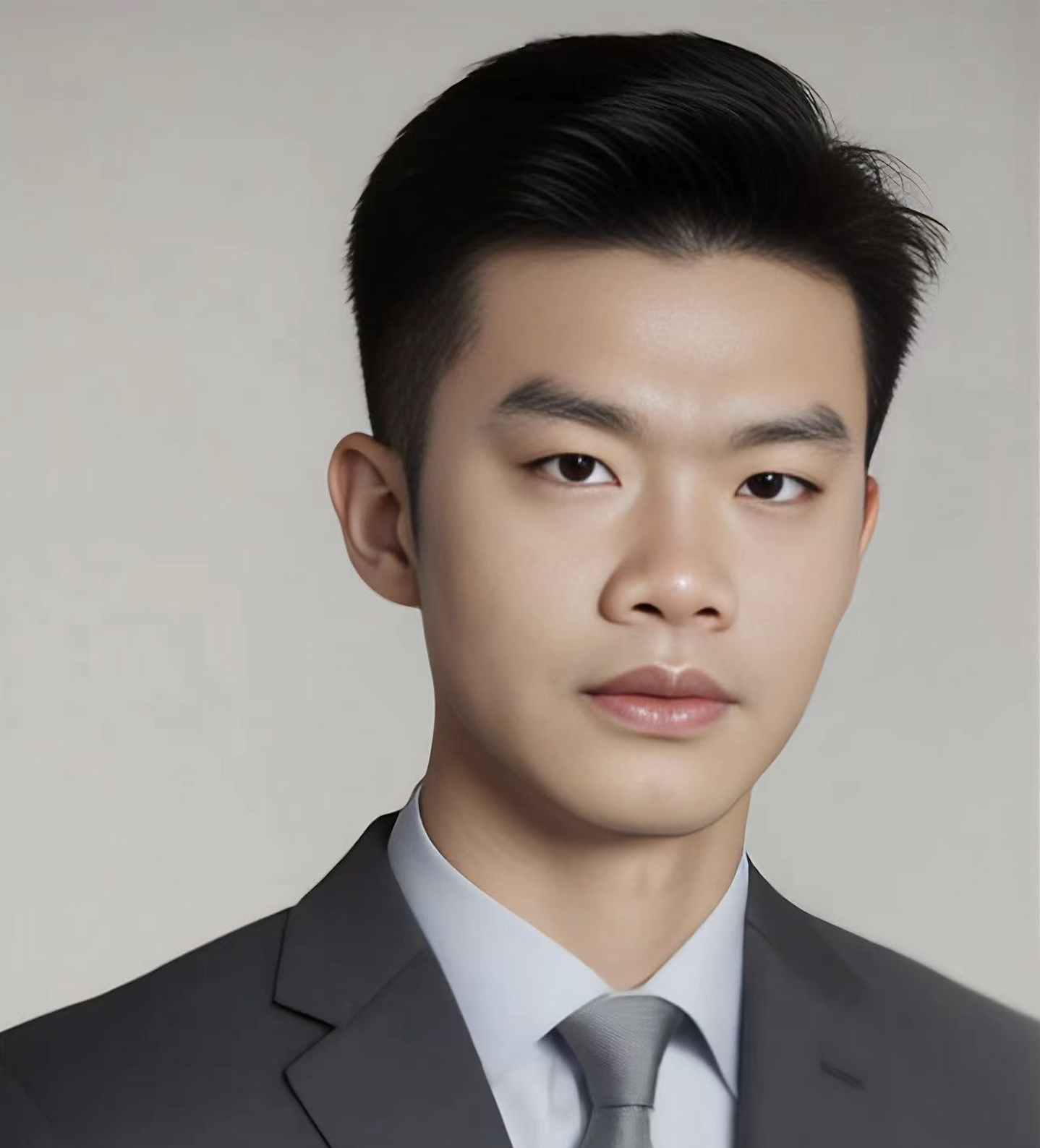}}]{Hao Tang} is an Assistant Professor at Peking University, China. Previously, he held postdoctoral positions at both CMU, USA, and ETH Zürich, Switzerland. He earned a master's degree from Peking University, China, and a Ph.D. from the University of Trento, Italy. He was a visiting Ph.D. student at the University of Oxford, UK, and an intern at IIAI, UAE. 
His research interests include machine learning, computer vision, embodied AI, and AIGC (including LLM), as well as their applications in scientific domains.
\end{IEEEbiography}

\begin{IEEEbiography}[{\includegraphics[width=1in,height=1.25in,clip,keepaspectratio]{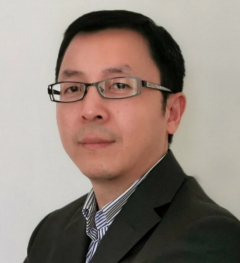}}]{Ling Shao} is a Distinguished Professor with the UCAS-Terminus AI Lab, University of Chinese Academy of Sciences, Beijing, China. He was the founder of the Inception Institute of Artificial Intelligence (IIAI) and the Mohamed bin Zayed University of Artificial Intelligence (MBZUAI), Abu Dhabi, UAE. His research interests include generative AI, vision and language, and AI for healthcare. He is a fellow of the IEEE, the IAPR, the BCS and the IET.
\end{IEEEbiography}

\begin{IEEEbiography}[{\includegraphics[width=1in,height=1.25in,clip,keepaspectratio]{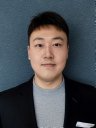}}]{Zhenyu Zhang}
	 is a staff research scientist at Tencent Youtu Lab, where he works on computer vision, graphics and machine learning. He got the Ph.D degree from Department of Computer Science and Engineering, Nanjing University of Science and Technology in 2020, supervised by Jian Yang. In 2019, He spent 10 months as a visiting student at MHUG group in Unviversity of Trento, Italy, supervised by Nicu Sebe. He is interested in 3D reconstruction, neural rendering and implicit neural representation. 
\end{IEEEbiography}

\begin{IEEEbiography}[{\includegraphics[width=1in,height=1.25in,clip,keepaspectratio]{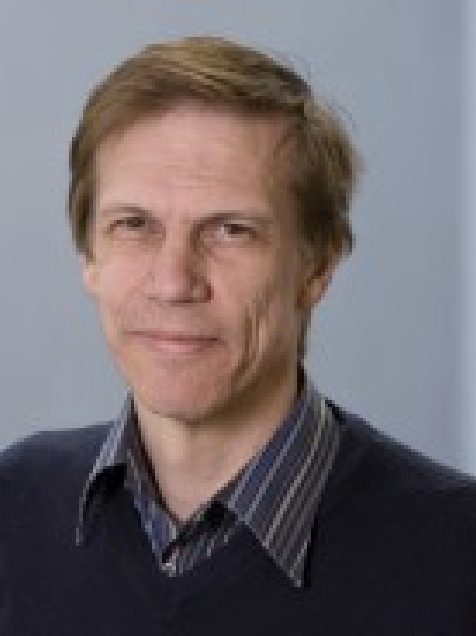}}]{Luc Van Gool} is full professor for computer vision at INSAIT, Sofia Un.
He is professor emeritus at ETH Zurich and the KU Leuven. He has been a
program committee member of several, major computer vision conferences
(e.g. Program Chair ICCV'05, Beijing, General Chair of ICCV'11,
Barcelona, and of ECCV'14, Zurich). His main interests include 3D
reconstruction and modeling, object recognition, and autonomous driving.
He is a co-founder of more than 10
spin-off companies. 
\end{IEEEbiography}

\begin{IEEEbiography}[{\includegraphics[width=1in,height=1.25in,clip,keepaspectratio]{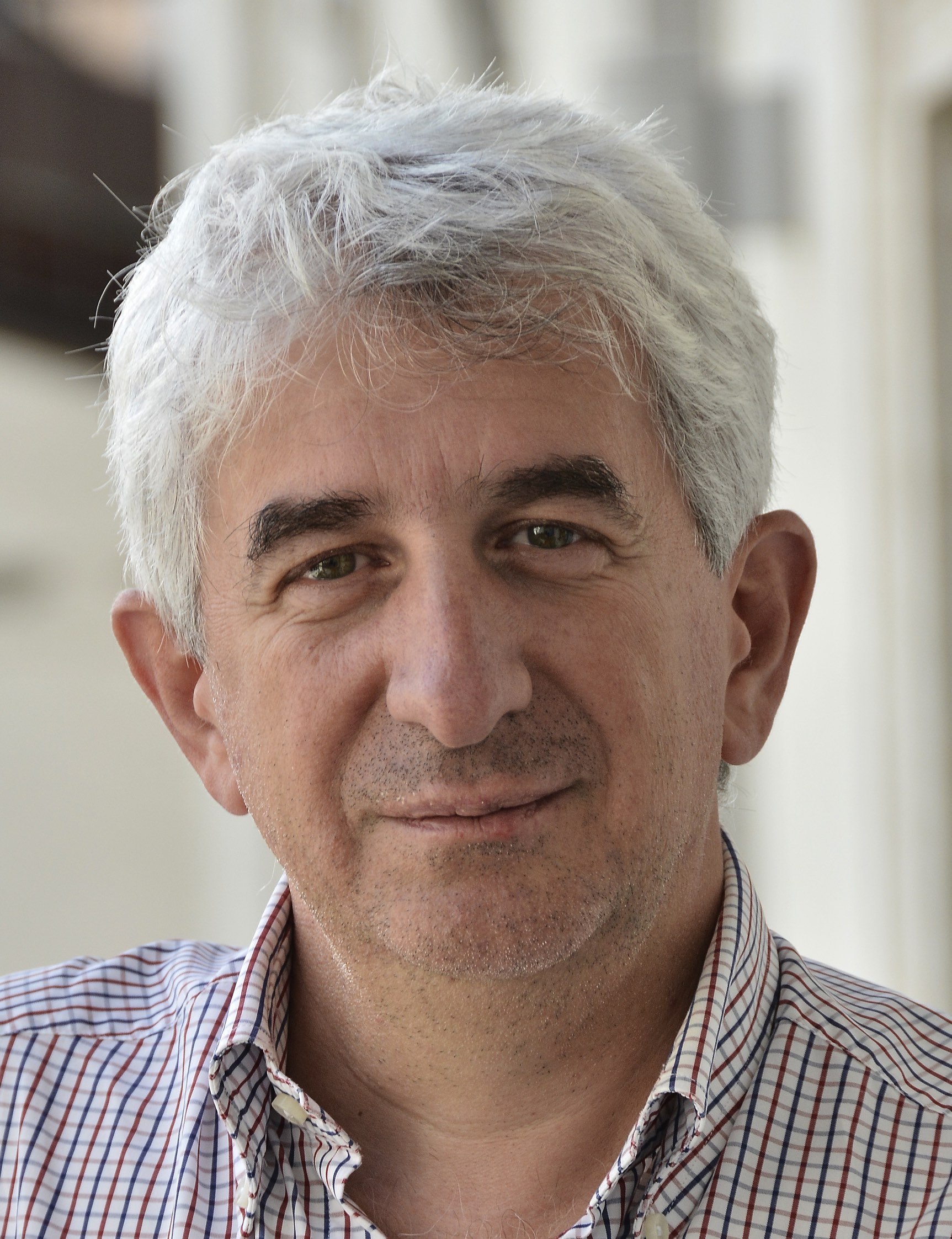}}]{Nicu Sebe}
is Professor with the University of Trento, Italy, leading the research in the areas of multimedia information retrieval and human behavior understanding. He was the General Co-Chair of ACM Multimedia
2013 and 2022, and the Program Chair of ACM Multimedia 2007 and 2011, ECCV 2016, ICCV 2017 and ICPR 2020. He is a fellow of ELLIS and of the International Association for Pattern Recognition (IAPR). He is the co-editor in chief of the journal of Computer Vision and Image Understanding.
\end{IEEEbiography}

\end{document}